\documentclass{article}

\usepackage[final]{corl_2024} % Uncomment for the camera-ready ``final'' version.
\usepackage{url}            % simple URL typesetting
\usepackage{booktabs}       % professional-quality tables
\usepackage{amsfonts}       % blackboard math symbols
\usepackage{nicefrac}       % compact symbols for 1/2, etc.
\usepackage{microtype}      % microtypography
\usepackage{graphicx,wrapfig,lipsum}
\usepackage{epsfig} % for postscript graphics files
\usepackage{times} % assumes new font selection scheme installed
\usepackage{amsmath} % assumes amsmath package installed
\usepackage{amssymb}  % assumes amsmath package installed
\usepackage{caption}
\usepackage{lipsum}
\usepackage{float}
\usepackage{algorithm}
\usepackage{algorithmic}
\usepackage{color}
\usepackage{booktabs}
\usepackage{subfigure}
\usepackage{subcaption}
\usepackage{setspace}
\usepackage{bm}

\usepackage{multirow}

\usepackage{hyperref}
\usepackage{multirow}
\usepackage{bm}
\usepackage{listings}
\usepackage{caption}
\usepackage{graphicx}
\usepackage{float} 
\usepackage{array}
\usepackage{colortbl}
\usepackage[labelfont=bf]{caption}
\usepackage{wrapfig}
\usepackage[utf8]{inputenc}

\newcommand{\XY}[1]{{\color{black} #1}}

\title{HiRT: Enhancing Robotic Control with Hierarchical Robot Transformers}

\author{
 Jianke Zhang\footnotemark[1]$^{~~1}$, Yanjiang Guo\footnotemark[1]$^{~~1}$, Xiaoyu Chen$^{1}$,\\ \textbf{Yen-Jen Wang$^{2}$}, \textbf{Yucheng Hu$^{1}$}, \textbf{Chengming Shi$^{1}$}, \textbf{Jianyu Chen\footnotemark[2]$^{~~1,3}$}\\
  $^{1}$Institute for Interdisciplinary Information Sciences, Tsinghua University\\
  $^{2}$University of California, Berkeley\\
  $^{3}$Shanghai Qizhi Institute\\
  \{zhangjk24, guoyj22, chen-xy21, huyc24, shicm19\}@mails.tsinghua.edu.cn,\\ wangyenjen@berkeley.edu, jianyuchen@tsinghua.edu.cn\\
}
\begin{document}
\maketitle

\renewcommand{\thefootnote}{\fnsymbol{footnote}} %将脚注符号设置为fnsymbol类型，即特殊符号表示
\footnotetext[1]{These authors contributed equally to this work.} %对应脚注[1]
\footnotetext[2]{Corresponding authors.} %对应脚注[2]
%===============================================================================
\begin{abstract}
%However, their deployment involves substantial computational costs and inference delays, resulting in the incapability of complex dynamic tasks.
% Previous methods, such as reinforcement learning and diffusion models, either lack efficiency or generalization capability. 
% Motivated by dual process theory, which differentiates between slow analytical planning and fast intuitive actions, this study proposes HiRT, a hierarchical imitation learning framework. HiRT leverages VLMs for long-term planning at low frequency and small policy models for quick high-frequency responses, enhancing both multi-task generalization and inference speed. 
% Vision-language models (VLMs) have demonstrated remarkable abilities in understanding visual and linguistic information, representing strong generalization capabilities when applied to embodied agents through task-level planning or vision-language-action (VLA) fine-tuning. 
% However, large VLA policy resulted in low-frequency execution and thus are limited to quasi-static tasks and struggles with dynamic tasks that require rapid interaction with environments.
% due to their imitation learning approach in training on quasi-static tasks, combined with high computational costs and inference latency, they do not perform well in dynamic tasks that require quick action execution. 
Large Vision-Language-Action (VLA) models, leveraging powerful pre-trained Vision-Language Models (VLMs) backends, have shown promise in robotic control due to their impressive generalization ability. However, the success comes at a cost. Their reliance on VLM backends with billions of parameters leads to high computational costs and inference latency, limiting the testing scenarios to mainly quasi-static tasks and hindering performance in dynamic tasks requiring rapid interactions. To address these limitations, this paper proposes \textbf{HiRT}, a \textbf{Hi}erarchical \textbf{R}obot \textbf{T}ransformer framework that enables flexible frequency and performance trade-off. 
HiRT keeps VLMs running at low frequencies to capture temporarily invariant features while enabling real-time interaction through a high-frequency vision-based policy guided by the slowly updated features. 
Experiment results in both simulation and real-world settings demonstrate significant improvements over baseline methods. Empirically, in static tasks, we double the control frequency and achieve comparable success rates. Additionally, on novel real-world dynamic manipulation tasks which are challenging for previous VLA models, HiRT improves the success rate from 48\% to 75\%.

\end{abstract}
% Two or three meaningful keywords should be added here
\keywords{Imitation Learning, Robots, Vision Language Models} 

%===============================================================================

\section{Introduction}

\XY{Large Vision-Language-Action (VLA) models  \cite{brohan2023rt,li2023vision}  provide a principled way to combine large vision-language models (VLMs) \cite{li2023blip,wang2022image,dai2024instructblip,driess2023palm} with end-to-end training on embodied tasks. Building on the top of pre-trained VLMs, existing VLA models \cite{brohan2023rt,li2023vision} propose to tune VLMs on massive robot data, which enables the direct end-to-end robot control while enjoying the benefits of VLM pretraining. 
Existing works mostly focus on multi-task generalization, enhancing performance in zero-shot and few-shot learning across various tasks.}

\XY{Though the VLM backends with billions of parameters bring superior generalization advantages, it comes at the cost of the heavy computational burden. During deployment, it results in low control inference speed and high latency.} This can slow robot movements and extend task completion times, impairing performance and safety in dynamic tasks like manipulating fast-moving objects in cluttered environments \cite{ha2022flingbot,saxena2024mrest}.
\XY{The control frequency limitations of large VLA models remain a significant obstacle to deploying these advanced models on real-world robots.}

\XY{Inspired by the dual process theory of human cognition~\citep{wason1974dual}, we propose HiRT, a hierarchical interactive imitation learning framework for VLA models. Dual process theory posits that there are two systems in human cognition: System 1, responsible for fast, intuitive reactions, and System 2, responsible for slow, analytical planning. Current VLA models can be seen as relying solely on System 2, using computationally expensive VLMs for inference and action generation. However, we argue that VLA models can benefit from merging these systems. HiRT utilizes System 2 to extract high-level, slowly changing information that guides a lightweight System 1 module. This System 1, implemented by a smaller model, can react swiftly to environmental changes. Though lightweight, System 1 in HiRT can leverage the guidance from System 2 to maintain performance comparable to the original VLM while obtaining notable speed gain.}

\begin{figure}[t]
    \centering
    \includegraphics[width=1.0\textwidth]{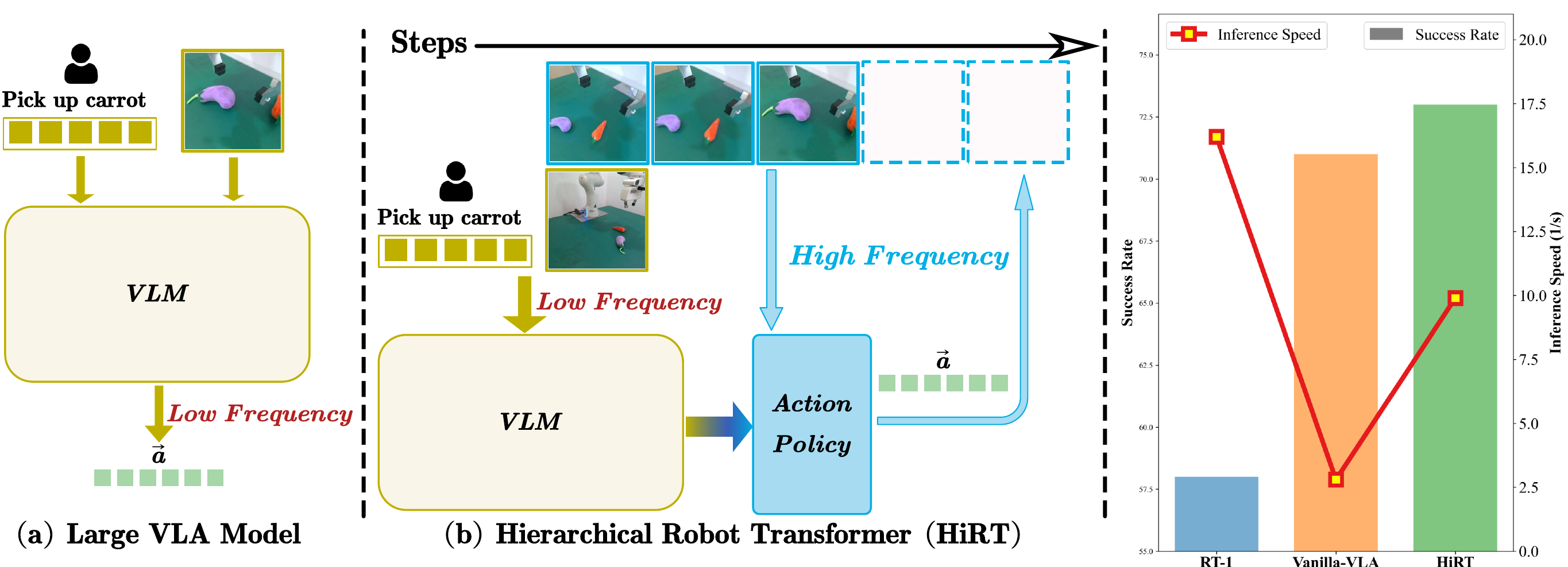}
    % \vspace{0.2mm}
    \caption{\textbf{Illustration of our proposed HiRT high-level architecture.} \textbf{(a)} Unlike large VLA models that directly output low-level actions with VLM, \textbf{(b)} HiRT is a hierarchical policy based on VLM. Given a task language instruction, the VLM encodes the observations into features that integrate multimodal information, and then a lightweight action policy conditions this latent to generate low-level actions asynchronously. 
    % HiRT enables asynchronous operation between the VLM and the action policy. 
    As shown in \textbf{(c)}, our method can achieve higher performance and significantly improve inference speed. 
    % \XY{plot y axis for speed and success rate,}
    }
    %与大型VLA模型（直接使用VLM输出底层动作）（a）的方法不同，HiRT是一种基于VLM的层次化策略（b）。给定一个任务语言指令，VLM将之前的视觉场景编码为融合了多模态信息的特征。我们提出的轻量级的动作策略以VLM的Latent作为条件，在每个时间步输出底层动作。HiRT可以使VLM与动作策略异步运行。如（c）所示，当VLM以不同的异步间隔运行时，我们的方法可以在调整性能与推理速度：取得相较于VLA方法更高的性能或以微弱的性能损失换取巨大的速度提升。
    \label{overview}
    \vspace{-1mm}
\end{figure}

We term this approach HiRT, a hierarchical interactive imitation learning framework designed for rapid execution across a variety of instructions, scenes, and tasks. HiRT consists of two primary components: the understanding module and the execution module. The understanding module, InstructBLIP (7B)~\citep{dai2024instructblip}, is a pre-trained large visual-language model that transforms visual information and language instructions into latent features enriched with commonsense knowledge for long-term scene understanding, including task planning and error correction. The execution module is a compact visual-based action policy that processes short-term scene cognition, utilizing prior observations and latent features from the visual-language model. To enhance focus on global instruction and visual data, we incorporate novel conditioning layers within the execution module. HiRT leverages the slower visual-language model to guide the swift low-level policy, enabling efficient performance in both quasi-static and dynamic tasks at high frequencies. Additionally, we achieve further speed optimizations by adjusting the asynchronous frequency of the modules.

\section{Related Works}
\textbf{Language-Conditioned Imitation Learning for Robot Manipulation.}
The study of integrating language with robotic actions \citep{macmahon2006walk,tellex2011understanding,tellex2020robots} through imitation learning has a long history, where language is commonly used as goal specification \citep{brohan2022rt,stone2023open,jang2022bc,thomason2020jointly} or intermediate representation for planning \citep{li2023camel,ma2023liv,nair2022r3m}. Some prior works have employed reinforcement learning techniques \citep{nair2022learning,luketina2019survey,misra2017mapping,jiang2019language,goyal2021pixl2r} to solve certain types of downstream tasks. To address incapability in generalization of these RL methods, recent works concentrate on prompting Large Language Models (LLMs)~\citep{li2023camel,oh2017zero,andreas2017modular,ahn2022can,crispino2023agent} for high-level task planning and fine-tuning vision-language models (VLMs) on expert robotic datasets for low-level robotic control~\citep{nair2022learning,brohan2022rt,goyal2021pixl2r,stepputtis2020language,shridhar2022cliport,mei2016listen}.
% supervised behavior cloning methods \citep{jang2022bc,brohan2022rt,stepputtis2020language,shridhar2022cliport,mei2016listen}. Among these approaches, some existing works concentrate on solving long-horizon tasks with high-level instructions \citep{li2023camel,oh2017zero,andreas2017modular,ahn2022can,crispino2023agent}, while others use end-to-end manners to address low-level instructions \citep{nair2022learning,brohan2022rt,goyal2021pixl2r,stepputtis2020language}. 
% Like the latter works, we focus on solving various manipulation tasks through end-to-end imitation learning. 
Different from previous works that explore how to generalize to new tasks,
%instructions \citep{nair2022learning,lynch2020grounding,hill2020human} and tasks \citep{jang2022bc,brohan2022rt},
% While previous works have explored how to enhance policy generalization to new instructions \citep{nair2022learning,lynch2020grounding,hill2020human} and tasks \citep{jang2022bc,brohan2022rt} through pre-trained language embeddings, 
we focus on solving low-level manipulation tasks by leveraging the extensive visual-linguistic knowledge within VLMs more efficiently and effectively.

\textbf{Vision-Language Models for Robotics.}
Applying pre-trained VLMs ~\citep{li2023blip,wang2022image,dai2024instructblip,driess2023palm,wang2023prompt} to various embodied scenarios is a recent focal area of research. Most of the prior works focus on using VLMs for high-level planning or reasoning~\citep{ahn2022can,zeng2022socratic,shah2023lm,huang2022inner,huang2023visual,song2023llm,liu2023llm+}. To effectively connect visual or linguistic information with the physical environment, embodied models need to fine-tune pre-trained VLMs on embodied data~\citep{brohan2023rt} including video data containing task-level planning in linguistic form~\citep{mu2024embodiedgpt,li2023camel,ahn2022can}, simple text descriptions~\citep{wu2023unleashing,bahl2023affordances}, low-level actions~\citep{belkhale2024rt,gu2023robotic,shah2023vint} (known as vision-language-action models).
%~\citet{brohan2023rt} showed that pre-trained VLMs possess not only multimodal reasoning capabilities but also the ability to bridge the gap between visual-linguistic information with low-level robotic actions. Subsequent research has explored methods such as reinforcement learning~\citep{szot2023large,carta2023grounding} and the use of human first-person operation videos~\citep{wu2023unleashing} to address the scarcity of embodied data.
%Despite the strong generalization capabilities of large embodied foundation models based on VLMs to new visual scenes, tasks, and instructions, 
However, deploying such large VLA models often results in slow inference speeds~\citep{hu2023toward}, which makes embodied models unsuitable for scenarios requiring precise operations or quick execution. Our approach focuses on addressing this limitation by using a novel policy model, which can effectively retain the robust visual-linguistic capabilities of the larger models.
% 将预训练的视觉语言模型应用于各种具身场景是近期研究的热点，其中大部分工作集中于利用VLM进行高层规划或推理。为了利用VLM构建视觉或语言信息与物理环境的连接，带有VLM的具身模型需要学会同时利用互联网图像文本数据和具身数据，后者主要包括带有底层动作信息的机器人轨迹、含有语言形式高层规划信息或者简单文本描述的视频数据。RT2工作中展示了预训练的VLM不仅具有多模态推理能力，还展示出了将视觉语言信息落地在机器人底层空中的能力。后续的一些工作探索了利用强化学习和人类第一视角操作视频GR1的方法解决将VLM应用在策略学习中需要大量稀缺的具身数据的问题。尽管基于VLM构建的大型具身基础模型有很强泛化到新的视觉场景、新任务、新指令的能力，但是大型模型直接部署在真实机器人时会导致过慢的推理速度，这使得具身模型不能很好地用于需要精细操作或快速执行的场景中。我们关注于通过小的语言条件化策略模型提高大模型在推理时的效率，兼顾大模型的视觉语言理解能力。

\textbf{Hierarchical Action Planning.}
Hierarchical action planning~\citep{li2023camel,du2023video,ahn2022can,abeyruwan2023agile,guo2023doremi} involves decomposing a task into multiple simpler tasks that can be executed directly, enabling strategies to tackle more complex, long-horizon tasks. 
Previous works have demonstrated the role of inputting prompts into LLMs as a bridge to low-level actions. Specifically, this can be implemented through task-level planning~\citep{lin2023text2motion,mu2024embodiedgpt,du2023video}, code execution~\citep{singh2023progprompt,liang2023code,wang2023gensim}, or other planning representations such as 3D scene graph~\citep{rana2023sayplan}, affordance function~\citep{huang2023voxposer}, and action pattern for locomotion~\citep{tang2023saytap}. 
However, these approaches are typically agnostic to physical embodiment, preventing the high-level models from directly interacting with the physical environment. 
In contrast to these methods, we ground VLMs to a specific robot's physical form in an end-to-end manner, enabling it to learn hierarchical task planning through continuous intermediate representations. 
% Our approach focuses on leveraging hierarchical model structures to enhance the reasoning speed of the model. Empirically, we demonstrate that this method balances task performance and execution efficiency.
% 层次化的动作规划是指将任务按照复杂程度分解为多个可以直接执行的简单任务，使策略能够完成较为复杂的long-horizon任务。先前的工作向大语言模型或视觉语言模型提供语言提示词进行任务规划问答，展示出了大模型通过少样本微调后在物理环境中进行规划、编写执行代码或与人类通过语言交互的能力。然而，这些工作一般采用了训练好的底层策略执行上层给出的任务规划，导致上层大模型无法直接与物理环境进行交互。与这些尝试分解高层任务的方法不同，我们在训练时以端到端的方式使模型能够通过连续的中间表示学习不同层次的任务规划，同时我们专注于利用层次化的模型结构提升模型的推理速度。实验上我们展示出了这种方法能够取得任务性能和执行效率间的平衡。

\section{Method}

In this section, we first establish the problem in Sec.\ref{prob}. Then, we present HiRT, a hierarchical policy architecture that supports multi-task learning and fast inference in Sec.\ref{backbone}. The key intuition is to draw help from pre-trained VLMs to extract rich semantic representations from multi-modal inputs, and then apply these representations to lightweight action policies that can operate asynchronously and independently of the VLM. 
% This requires its capability for visual-language alignment, long-term planning, and multi-step scene understanding. 
Specifically, HiRT explores a popular vision-language model, InstructBLIP~\citep{dai2024instructblip}, utilizing its open-source model as the backbone.
We aim to output low-level actions with a latent-conditioned policy that leverages historical observations and latent encoded by VLM. This small-scale policy should operate independently of the large model at a higher frequency, necessitating a compact architecture composed of lightweight visual encoder. Following BC-Z~\citep{jang2022bc} and RT-1~\citep{brohan2022rt}, we design a latent-conditioned model as the lower-level policy, capable of independently performing behavior cloning for a limited number of tasks at high frequency.
% Finally, we describe the experimental setup, including the optimization objectives and the task settings for both the simulation and real-world environments.

\subsection{Problem Formulation and Method Overview}\label{prob}
The language-conditioned manipulation problem can be considered a decision sequence under the environment modeled by Markov decision process: $(S, A, R, P, \rho_0)$, where $S, A, \rho_0$ represents state space, action space and initial state distribution respectively, $R: S \times A \times S \rightarrow \mathbb{R}$ represents the reward function, indicating whether a wanted state or task has been completed, $P: S \times A \times S \rightarrow [0, 1]$ represents probabilistic forward dynamics function of the environment. 
Specifically, given a free-form language instruction $l$ specifying a certain task, the control policy receives a visual observation \( \bm{o}\) which is typically composed of a series of images. Then an action \(\bm{a}\in A\), incorporating the relative position and pose of the end effector, is sampled from an action distribution  \( \pi(\cdot|\bm{o},l) \) modeled by control policy. 

For HiRT, the policy \( \pi(\bm{a}|\bm{o},l) \) is parameterized by 
\( F_\theta \) from the vision language model and \( S_\phi \) from the swift latent-conditioned policy. 
At certain time steps in the trajectory $\hat t_k\in\{t_i\}_{i=1}^T, k\leq T$, the VLM backbone takes in a visual observation \mbox{$\bm{\bar{o}}_{\hat t_k}=Sample(\bm{o}_{:\hat t_k})$} obtained through asynchronous sampling and a natural language instruction \(l\), and outputs a fused embedding: \mbox{$\bm{z}_{\hat t_k}=F_{\theta}(\bm{\bar{o}}_{\hat t_k},l)$}. Simultaneously, at each time of step, the latent-conditioned model predicts actions with recent context of visual observations and the latest latent: $\bm{a_t}=S_\phi(\bm{o}_{:t},\bm{z}_{\hat t_k})$. The specific details of the modules will be explained in the following section.
\subsection{The HiRT Framework}\label{backbone}

\begin{figure}[t]
    \centering
    \includegraphics[width=1.0\textwidth]{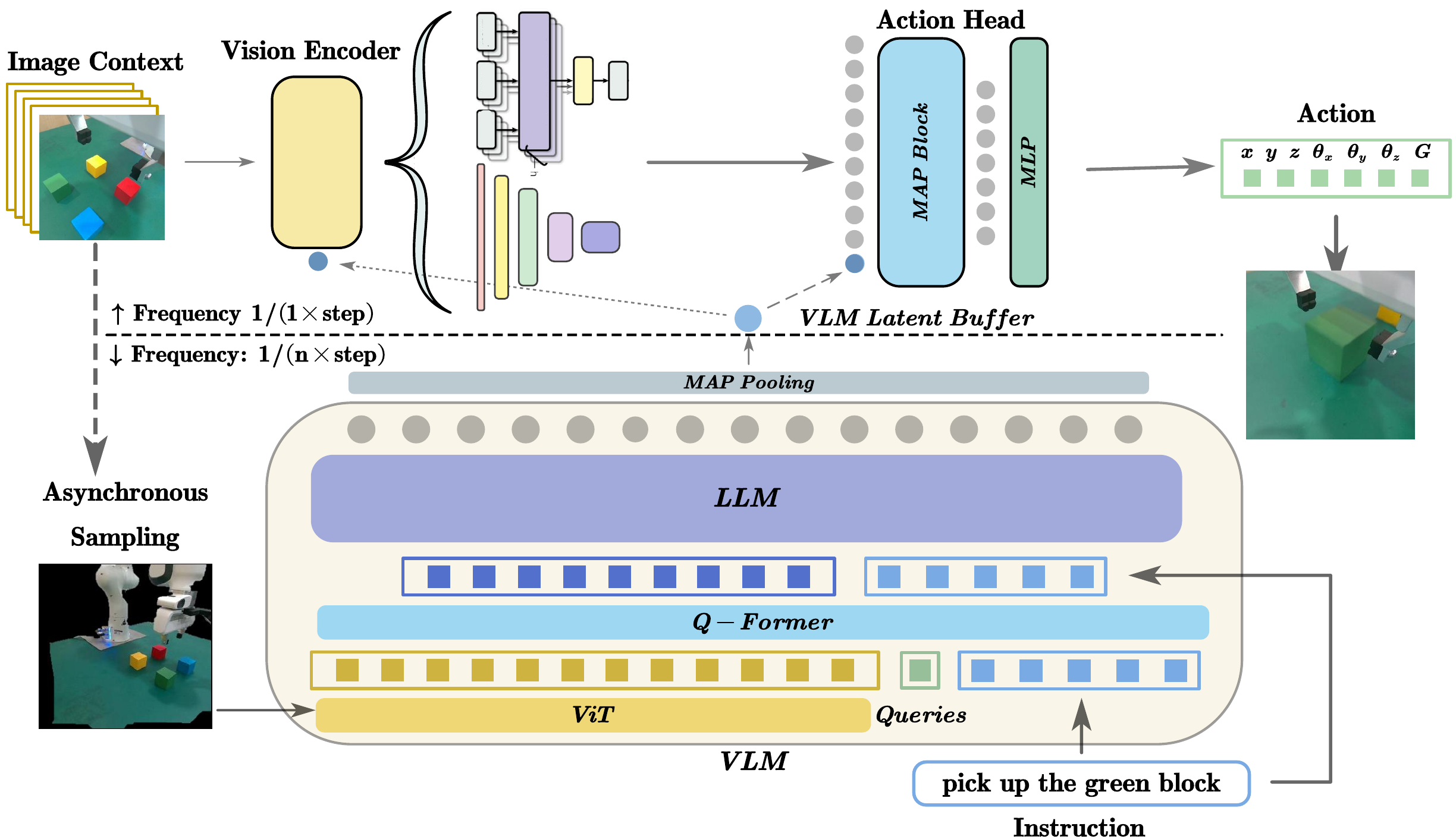}
    % \vspace{0.2mm}
    \caption{\textbf{HiRT network architecture.} The instruction is transformed into a continuous latent with sampled visual observation with a vision-language model and is cached into a latent buffer. At each step of inference, the pre-trained vision encoder encodes visual observations conditioned on the latest latent, and then the reduced vision-language tokens are decoded to low-level action with a conditioned action head.}
    \label{method}
    \vspace{-1mm}
\end{figure}

\subsubsection{Encoding Multi-modal Information with Vision-Language Model}
%在HiRT中InstructBLIP利用单个图片形式的视觉信号$\bm{\bar{o}}$对指令$l$进行编码。InstructBLIP中由预训练的视觉编码器、大语言模型和可学习的Querytoken以及Q-former组成。在每一个执行VLM的时间步$t_k$,相机图片（手腕相机或第三视角相机）被Vision Transformer （VIT）编码为视觉令牌序列$\hat X_{t_k}^o=ViT(\bm{\bar{o}}_{\hat t_k}) \in \mathcal R^{N\times d}$,其中N表示令牌长度，d表示令牌宽度。随后，$\hat X_{t_k}^o$与指令令牌$X_{t_k}^l$和可学习的查询令牌拼接，被Q-Former(一个轻量级transformer）编码为融合了语义信息的图片表征: $X_{t_k}^o=QFormer(\hat X_{t_k}^o,X_{t_k}^t)$。最后视觉查询特征作为指令文本的提示词一同输入到大语言模型（LLaMA）中。令大语言模型中第l层嵌入${X^l}_{t_k}=({x_1}^l,{x_2}^l,\cdots,{x_N}^l)_{t_k}$,l层输出${X^{l+1}}_{t_k}$被如下计算：
% \begin{equation}
% \begin{align}
%     &\hat X^l_{t_k}=MSA(LN(X^l_{t_k}))+X^l_{t_k},~l=1,\cdots,L\\
%     &X^{l+1}_{t_k}=MLP(LN({\hat X^l}_{t_k}))+{\hat X^l}_{t_k}
% \end{align}
%其中$X^1_{t_k}=(X_{t_k}^o,X_{t_k}^l)$,L表示LLM中transformer的深度，MSA表示多头注意力模块，MLP表示多层感知机，LN表示LayerNorm。不使用LLM最后一层输出$X^{L+1}_{t_k}=({x_{1}^{L+1},x_2^{L+1},\cdots,x_N^{L+1})_{t_k}$作为语言令牌的生成，我们希望借助这一informative的语言嵌入用于指示动作的生成。为了实现这一目的，我们利用MAP模块聚合这一表征:$\bm{x}_{t_k}=MAP(X^{L+1}_{t_k})$，并将其暂存到VLM-Latent Buffer指导动作策略。
% \end{equation}
In HiRT, InstructBLIP~\citep{dai2024instructblip} encodes the instruction \( l \) using a visual signal \(\bm{\bar{o}}\) in the form of a single image. InstructBLIP comprises a pretrained visual encoder, a large language model (LLM), learnable query tokens, and a Q-Former~\citep{li2023blip}. At each execution time step \(\hat t_k \), the visual observation (from either the wrist or third-view camera) is encoded by a Vision Transformer (ViT)~\citep{dosovitskiy2020image} into a sequence of visual tokens:
\[\hat{X}_{t_k}^o = ViT(\bm{\bar{o}}_{t_k}) \in \mathbb{R}^{N \times d}\]
where \( N \) denotes the token length and \( d \) the token width.
Subsequently, \(\hat{X}_{t_k}^o\) is concatenated with the instruction tokens \( X_{t_k}^l \) and learnable query tokens $X^Q$, and encoded by the Q-Former (a lightweight transformer) into an image representation fused with semantic information: 
\[X_{t_k}^o = QFormer(\hat{X}_{t_k}^o, X_{t_k}^l,X^Q) \]\
Finally, these visual query features are used as prompts for the pre-trained LLM (LLaMA~\citep{touvron2023llama}). Set the embeddings at layer \( i \) as \(X^i_{t_k}\), the output at layer \( i+1 \) is computed as follows:
\begin{align}
    &\hat{X}^i_{t_k} = MSA(LN(X^i_{t_k})) + X^i_{t_k}\\
    &X^{i+1}_{t_k} = MLP(LN(\hat{X}^i_{t_k})) + \hat{X}^i_{t_k}\\
    X^1_{t_k} = (X_{t_k}^o, X^l_{t_k}&), X^i_{t_k} = (x_1^i, x_2^i, \cdots, x_N^i)_{t_k}, i = 1, \cdots, L 
\end{align}
where \( L \) denotes depth of transformer layers in the LLM, \(MSA\) represents the multi-head attention module, \(MLP\) stands for the multi-layer perceptron, and \(LN\) denotes LayerNorm. Instead of generating language tokens from the final layer output \mbox{\(X^{L+1}_{t_k}\)}, we aim to use the informative language embeddings to guide action generation. We employ a MAP module~\citep{pmlr-v97-lee19d}, a single layer of attention block, to aggregate these representations: \(\bm{x}_{t_k} = MAP(X^{L+1}_{t_k})\), which will be used for conditioning the action policy in Sec.\ref{lcp}.

\subsubsection{Latent-Conditioned Policy}\label{lcp}
%为了实现高频的动作控制，底层的策略必须是轻量级的网络。跟随同样利用指令和视频作为task embedding生成动作的bcz framework，我们使用EfficientNet将图像上下文$\bm{o}_{:t}$编码为视觉特征。这些视觉特征被VLMlatent条件化，通过FiLM layers。同时，我们考虑引入RT1中的Transformer。为了进一步融合第一视角的视觉信息与VLM-Latent-Buffer中的informative embedding，Transformer中的每一层嵌入都会被Latent$\bm{x}_{t_k}$条件化。最后，我们使用MAPblock聚合Transformer的全部视觉表征，将其映射到连续的动作空间。
% The low-level policy must be a lightweight network to achieve high-frequency action control. 
Following the BC-Z~\citep{jang2022bc} and RT-1~\citep{brohan2022rt}, which uses instructions and video as task embeddings, we encode the image context \(\bm{o}_{:t}\) into visual tokens $X^v_{:t}$  using a lightweight visual encoder, i.e., EfficientNet~\citep{tan2019efficientnet} and Vision Transformer~\citep{wang2022image}. Then, we use a MAP block to aggregate all the tokens into the continuous action space.
To further integrate the informative task embeddings encoded by VLM, we make use of the following conditioning strategies on either the visual encoder or action head:

\textbf{FiLM-Condition.} For visual encoder based on convolutional network (CNN), each hidden layers are conditioned on the VLM latent variable \(\bm{x}_{t_k}\). In EfficientNet, We use FiLM layers to compute the conditioned features:
\(
\hat{H} = FiLM(H \mid \bm{x}_{t_k}) = W_\gamma\bm{x}_{t_k}\cdot H + W_\beta\bm{x}_{t_k}
\), where \(H\) represents the hidden features, and \(W_\gamma,W_\beta\) are the learnable parameters in the FiLM layer. 

\textbf{Condition with Cross-Attention Layers.} 
In each self-attention layers of Transformer, we insert an additional cross-attention layer for conditioning:
\(
\hat{H} = CrossAttn(H, W_h\bm{x}_{t_k}) + H
\),
where \(W_h\) represents a learnable parameter that projects $\bm{x}_{t_k}$ to the space of hidden tokens \(H\).

\textbf{Condition with Prefix Tuning.} 
 To better enable VLM to regulate low-level actions, we utilize the VLM latent variable $x_{t_k}$ as a prefix prompt for the MAP block in the action head. Specifically, the actions are computed by $\bm a_t=MLP(MAP([x_{t_k},X^v_{:t}]))$.
% \textbf{Conditioning Strategy.}
%在latent-conditioned policy中，主干网络中的隐藏层都被VLM latent\(\bm{x}_{t_k}\)条件化。对于视觉编码器EfficientNet，我们利用FiLM layers来计算条件化的特征：
% \[\hat H=FiLM(H|\bm{x}_{t_k})=\gamma H+\beta,\gamma=W_\gamma(\bm{x}_{t_k}),\beta=W_\beta(\bm{x}_{t_k}) \]
%其中H表示视觉编码器中的隐藏特征，$W_\beta，W_\beta$分别表示FiLM layer中可学习的参数。Transformer中使用额外的交叉注意力层进行条件化，以进一步加强任务对场景和指令的理解：
% \[\hat H=CrossAttn(H,\bm{x}_{t_k})+H\]

\subsection{Training and Inference Strategy}
% During the inference phase, we can accelerate the model by adjusting the VLM's execution frequency. Specifically, at the initial time step \( t=0 \), the VLM encodes multi-modal information and stores it in a cache. The latent-conditioned policy reads the latent variable and quickly outputs actions. In the subsequent steps, the latent-conditioned policy uses the latest latent variable from the cache. At the same time, the VLM stops or runs asynchronously in parallel with the latent-conditioned policy. This asynchronous mechanism allows the policy to operate at nearly the same speed as the latent-conditioned policy, avoiding slow inference of the VLM.
\textbf{Asynchronous Operation and Sampling.}
%异步运行会导致latent-conditioned-policy使用包含多步之前场景和指令信息的latent。因此我们在训练阶段调整从历史观察中采样的输入给VLM的视觉信息形式来提升latent-conditioned-policy对latent的鲁棒性。考虑到第三视角信息更适合用于高层次规划，HiRT会在输入到latent-conditioned-policy中的第一视角观察$\bm{o}_{:t})$中随机选择一步，并将对应时间步的第三视角图片作为VLM的视觉输入$\bm{\bar{o}}_{\hat t_k}$。
During the inference phase, we can accelerate the model by adjusting the execution frequency of the VLM. Specifically, at the initial time step $t=0$, the VLM encodes multi-modal information with visual contexts and stores it in a cache.
% The latent-conditioned policy then reads the latent variable from the cache and quickly outputs actions. 
In subsequent steps, the latent-conditioned policy use the most recent latent variable from the cache to quickly output actions while the VLM runs asynchronously in parallel with the latent-conditioned policy. This asynchronous mechanism allows the policy to operate at nearly the same speed as the latent-conditioned policy, avoiding delays due to the VLM's slower inference. However, the asynchronous operation may cause the policy to use latent variables that reflect scene and instruction information from several steps earlier, which is misaligned with signals used in training. Therefore, during training stage, HiRT randomly selects a step from the past observation contexts \(\bm{o}_{:t}\) and uses the corresponding third-view image as the VLM's visual input. This technique can enhance robustness of the policy to the time-inconsistant latent variable. 

\textbf{Training Objective} During training, the VLM part is finetuned with LoRA ~\citep{hu2021lora} while rest of the network is fully finetuned.
Concretely, we utilize maximum likelihood imitation learning objectives. The desired relative position \(\bm a^{pos}\) of the end-effector (or continuous joint action) is optimized via regression loss (e.g. MSE loss). The discrete status \(a^{end}\) of the end-effector is optimized with binary cross-entropy loss:
\[\mathcal L=\sum_\mathcal B\frac{1}{|\mathcal B|}(||\bm a^{pos}-\bm\hat a^{pos}||_2^2+BCE(a^{end},\hat a^{end}))\]
where \(\bm \hat a^{pos},\hat a^{end}\) denote the demonstration for relative position and status of the end-effector in a sampled mini-batch \(\mathcal B\).
% \textbf{Accelerating via Asynchronous Mechanism}
%在推理阶段，我们可以通过调整调整VLM的运行间隔来加速模型。具体来说， 在初始$t=0$时VLM将多模态信息编码并存储在缓存，latent-conditioned-policy读取latent并快速输出动作。在之后的连续几步中，VLM停止latent-conditioned-policy每一步都使用缓存中最新的latent，独立于VLM运行。VLM停止或与latent-conditioned-policy并行异步地运行以更新最近场景的latent。通过这样的 异步机制可以使策略达到与latent-conditioned-policy完全相同的频率，避免了VLM前向传播很慢的影响。

\section{Experiments and Analysis}
\label{sec:result}
In this section, we conduct extensive experiments across three domains, including two simulated benchmarks Metaworld~\citep{yu2020meta} and Franka-Kitchen~\citep{gupta2019relay}, and a real-world panda manipulation environment to verify the effectiveness of our HiRT framework. We first introduce experiment setups in Sec.\ref{setup}. Then we present a quantitative analysis of the performance on quasi-static tasks, evaluating HiRT's capability to enhance inference speed while preserving generalization performance in Sec.\ref{staticexp}. Additionally, we test the performance in real-world dynamic tasks in Sec.\ref{dynamicexp}. Finally, we discuss design choices for implementing HiRT and perform ablation studies on key modules in Sec.\ref{ablation}. 
\subsection{Experiment Setup}\label{setup}
\textbf{Simulation Setup.}
%我们在Metaworld和Franka-Kitchen进行仿真实验。Metaworld BenchMark提供了50种独立的桌面操作任务。我们利用其中20种任务（每种包含50条专家数据）进行多任务学习。Franka-Kitchen中包含5种厨房操作任务。Following \textbf{infer}，在其中我们使用每种任务的100条专家数据进行多任务学习，并在额外两个新的场景测试泛化能力。
% We conduct simulated experiments on Metaworld and Franka-Kitchen. 
The Metaworld benchmark provides 50 distinct tabletop manipulation tasks, in which we use 20 tasks (each with 50 expert demonstrations) for multi-task learning. Franka-Kitchen includes 5 kitchen manipulation tasks. Following \citet{nair2022r3m}, we train policy models on 100 expert demonstrations for each task and test on tasks in origin and two new scenarios (alter the color scheme of the scene). We record the success rate to assess task performance: 20 attempts for each task in Metaworld and 100 for each task in Franka-Kitchen. To evaluate inference speed, we directly measure average time the policy takes to process 100 frames (avoiding influence of rendering). 
% 这块补一张静态和动态task的环境图

\textbf{Real World Setup.}
Our real-world experiments involve multiple quasi-static manipulation tasks on the Franka Emika Panda robot, involving picking and placing various objects, routing cables, pressing buttons, and opening drawers. Specifically, we collect 2000 trajectories including image observations from wrist and third-view cameras. For quasi-static tests, we place many other objects on the table to introduce distractions and we also test whether the model can grasp entirely new objects it has never seen before to verify its semantic grounding capabilities. Besides, we test the policy's performance on dynamic tasks by moving the target object at a roughly constant speed while the robotic arm executes its actions. All tasks involve randomization (e.g. the object's position, type, number of distracting objects, and the initial state of the gripper). We report success rate of each task over 20 attempts and the average time cost during real-world roll-out. More details on the design of experimental scenarios can be found in Appendix \ref{app_setup}, which can better demonstrate our testing of the generalization ability of semantic grounding in real scenarios.
\begin{figure}[t]
    \centering
    \includegraphics[width=1.0\textwidth]{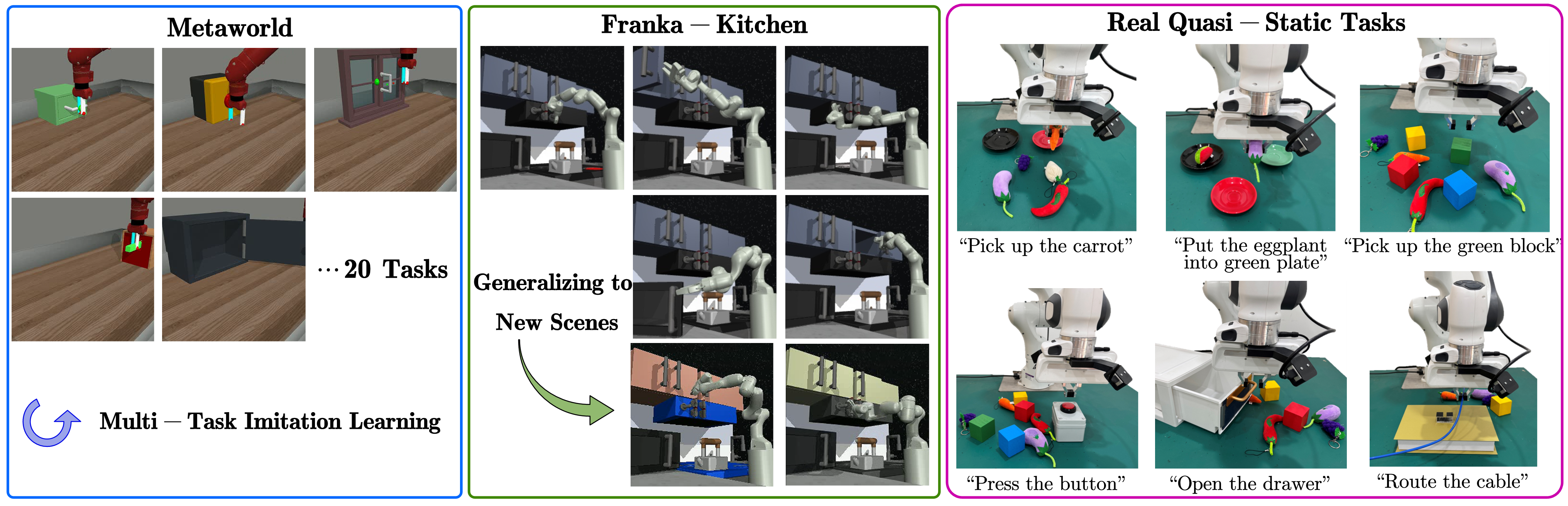}
    % \vspace{0.2mm}
    \caption{Visualization of the tasks in three domains. The left is Metaworld \cite{yu2020meta} in which we focus on the ability to learn multi-tasks. The middle depicts Franka-Kitchen~\citep{gupta2019relay} in which we study the ability to generalize to new scenes. The right shows our real-world settings, in which the model is trained on simple quasi-static tasks and tested on much more complex scenarios with unseen objects.}
    \label{fig: traintest}
    \vspace{-3mm}
\end{figure}
\subsection{Performance on static manipulation tasks}\label{staticexp}
%仿真环境的全部详细数据表ok
%真机全部数据表ok
%性能-速度比较图（每个方法4个bar：metaworld+franka_seen+franka_unseen+real_panda ok
%我们分别在如图\ref{trainset}所示的20个来自Metaworld的任务、Franka-Kitchen中5个任务以及Real-world中6类任务上训练HiRT，并在Seen和Unseen的任务进行测试。作为比较，我们还在相同环境下测试了Vanilla-VLA，我们使用VLM直接输出动作（可以被视为RT-2的复现）以及RT-1方法。
We train HiRT on 20 tasks from Metaworld, 5 tasks from Franka-Kitchen, and 4 skills from the real world (as shown in Figure \ref{fig: traintest}). For comparison, we evaluate Diffusion Policy (DP) ~\citep{chi2023diffusionpolicy}, Vanilla-VLA, which directly outputs actions from VLM (reimplementation of RT-2~\citep{brohan2023rt}), and RT-1~\citep{brohan2023rt} method under the same settings.
\begin{table}[t]
% \small
\centering
\resizebox{1.0\linewidth}{!}{
\begin{tabular}{*{1}{>{\centering\arraybackslash}p{2cm}}*{1}{>{\centering\arraybackslash}p{1.2cm}}|*{1}{>{\centering\arraybackslash}p{1.6cm}}|*{2}{>{\centering\arraybackslash}p{1.1cm}}|*{5}{>{\centering\arraybackslash}p{0.9cm}}}\toprule
  \multicolumn{2}{c|}{\textbf{Method}}&\textbf{Metaworld}&\multicolumn{2}{c|}{\textbf{Franka-Kitchen}}&\multicolumn{5}{c}{\textbf{Real-world}}\\\hline
  \textbf{Setting}&\textbf{Infer}&\textbf{Seen}&\textbf{Seen}&\textbf{Unseen}&\textbf{Pick-}& \textbf{Button-} & \textbf{Cable-}&\textbf{Drawer-}& \textbf{Average}\\
  \multicolumn{10}{c}{\vspace{-4.2ex}} \\  % 插入空行并缩短间距
  &\textbf{Speed/Hz}& \textbf{20 Tasks}&\textbf{5 Tasks}&\textbf{2 Tasks}& \textbf{Place} & \textbf{Press}  & \textbf{Route} & \textbf{Open}& \\
  \hline\hline
\textbf{RT-1}&20.1&65.8&63.8&33.0& 35.0 & 70.0 & 55.0 & 40.0 & 55.0\\
\textbf{DP}&4.6&52.2&-&-& 30.0 & 70.0 & 30.0 & 50.0& 45.0\\
\textbf{Vanilla-VLA}&4.1&73.8&73.4&70.0& \underline{70.0} & 85.0 & \underline{65.0} & \underline{65.0}& \underline{71.3}\\
\textbf{HiRT}&9.8&\underline{76.4}&\underline{80.8}&\underline{76.0}&\underline{70.0}& \underline{90.0} & 60.0 & 60.0 & 70.0\\
    \bottomrule
    \end{tabular}
% \begin{tabular}{ccccccccc}
% \toprule
% \textbf{Task}  &\textbf{Pick} &\textbf{Place} & \textbf{Button-} & \textbf{Cable-} &   \textbf{Drawer-} & \textbf{Drawer-} & \textbf{Average}&\textbf{Unseen-}\\
%         & \textbf{(4 tasks)}& \textbf{(3 tasks)} & \textbf{Press} & \textbf{Route}  & \textbf{Open} & \textbf{ Close}& \textbf{ Seen tasks}&\textbf{ tasks}\\
%  % & & & \textbf{-press} & \textbf{-press} & \textbf{-press} & \textbf{-press} & \textbf{-button} \\
% \midrule
% \textbf{Diffusion Policy}& 0.70 & 0.30 & 0.28 & 0.34& 0.42 & 0.58 &0.38\\
% \textbf{RT-1*}            & 0.72 & 0.52 & 0.40 & 0.34 & 0.44 & 0.50 & 0.43\\
% \textbf{Vanilla-VLA}            & \textbf{0.80} & 0.60 & 0.64 & 0.74 & \textbf{0.66} & 0.82 &0.69\\
% \textbf{HiRT}       & \textbf{0.80} & 0.55& 0.76   & 0.72 & 0.56 & 0.84 & 0.72\\ 
%     \bottomrule
% \end{tabular}
}
\vspace{2mm}
\caption{Success rates on quasi-static manipulation tasks.
%HiRT outperforms other baselines in both multi-task learning and generalization performance.
}
\vspace{-6mm}
\label{tab: result}
\end{table}

\textbf{Imitation and Zero-Shot Generalization Performance.}
% 表\ref{simresult}和表\ref{table_real}分别展示了仿真环境和真实环境的实验结果，其中HiRT在见过的任务和新任务中都有最高的成功率。使用了VLMLatent 进行条件化的HiRT相较于使用语言嵌入作为条件的RT-1在模仿的任务上成功率高20 percent,在新的任务场景中高30 percent。这reveal了VLM能够利用视觉场景提供更好的指令嵌入，帮助小型的动作策略泛化到新的任务中。另外，HiRT相较于Vanilla-VLA也有着显著的性能提升，这是因为小的latent-conditioned-policy可以关注历史视觉环境，并将融合的视觉特征ground到动作。
Table \ref{tab: result} presents the experimental results in simulation and real-world environments. HiRT achieves the highest success rates in simulated tasks and a high level of generalization capability similar to Vanilla-VLA in real-world environments. Compared to RT-1, which uses language embeddings for 
% \begin{wraptable}{htbp}{0.67\textwidth}
%   \centering
%   \begin{tabular}{*{1}{>{\centering\arraybackslash}p{2cm}}||*{1}{>{\centering\arraybackslash}p{1.8cm}}|*{2}{>{\centering\arraybackslash}p{1.9cm}}}\toprule
%   Method&Metaworld&\multicolumn{2}{c}{Franka-Kitchen}\\\hline
%   Setting&20 Tasks&Seen 5 Tasks&Unseen Tasks\\\hline\hline
% RT-1*~\citep{brohan2022rt}&-&63.8&33.0\\
% Vanilla-VLA&73.8&73.4&65.5\\
% HiRT&-&94.2&76.0\\
%     \bottomrule
%     \end{tabular}
%     % \vspace{2mm}
%       \caption{Success rates on simulation quasi-static manipulation tasks. HiRT outperforms the original method in both multi-tasking and generalization performance.}\label{simresult}
% \end{wraptable}
conditioning, HiRT, which utilizes vlm latent for conditioning, shows an average 20\% higher success rate on seen tasks and a 30\% higher success rate on new task scenarios. This demonstrates that VLM can leverage visual scenes to provide better instruction embeddings, aiding the small action policy in generalizing to new tasks. 
%Additionally, HiRT outperformed Vanilla-VLA significantly due to the smaller latent-conditioned policy's ability to focus on the historical visual environment and ground the integrated visual features into actions.

\begin{figure}[t]
\vspace{-2mm}
\begin{center}
\subfigure[Results in simulation tasks]{
\includegraphics[width=0.48\textwidth]{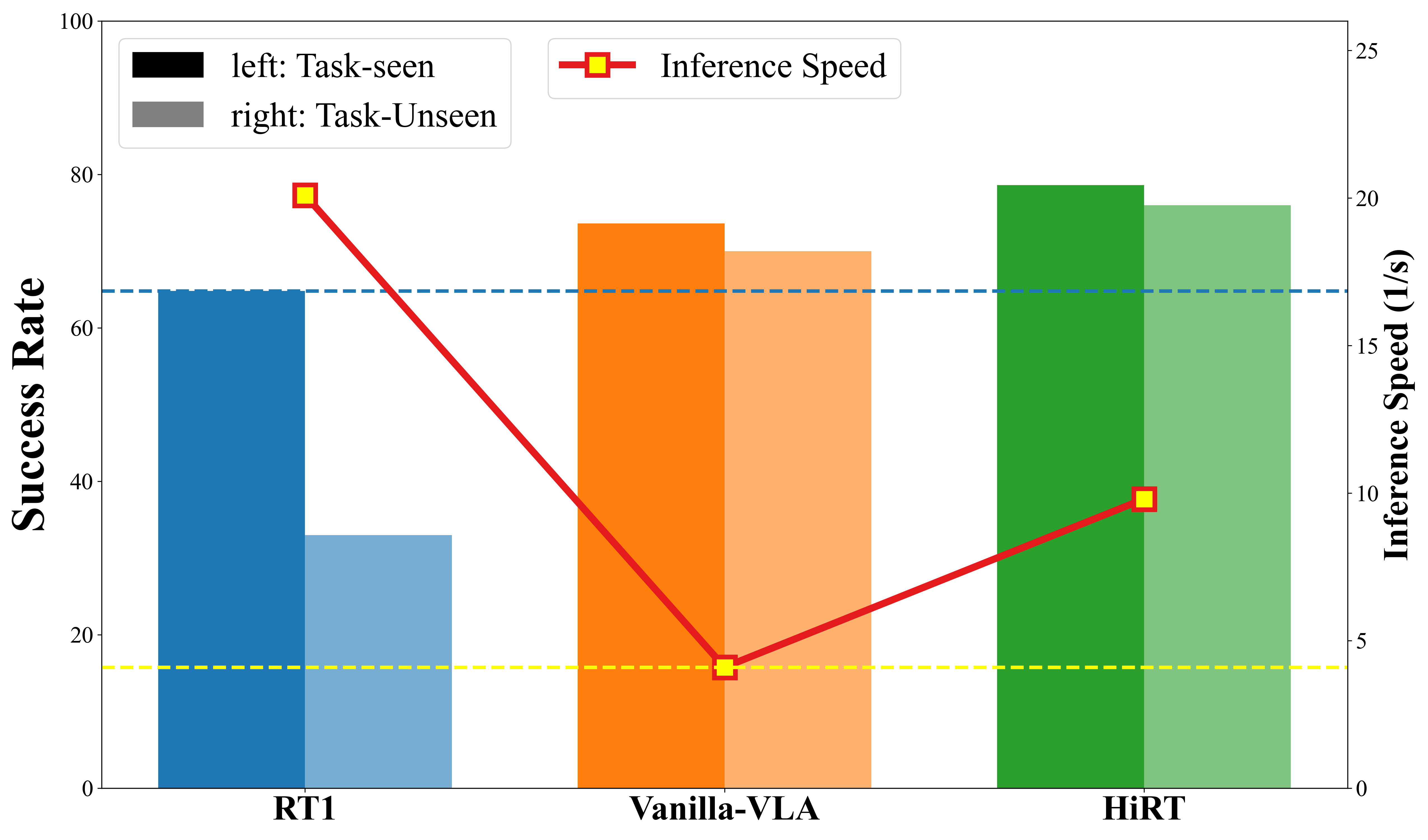}}
\subfigure[Results in real-world quasi-static tasks]{
\includegraphics[width=0.48\textwidth]{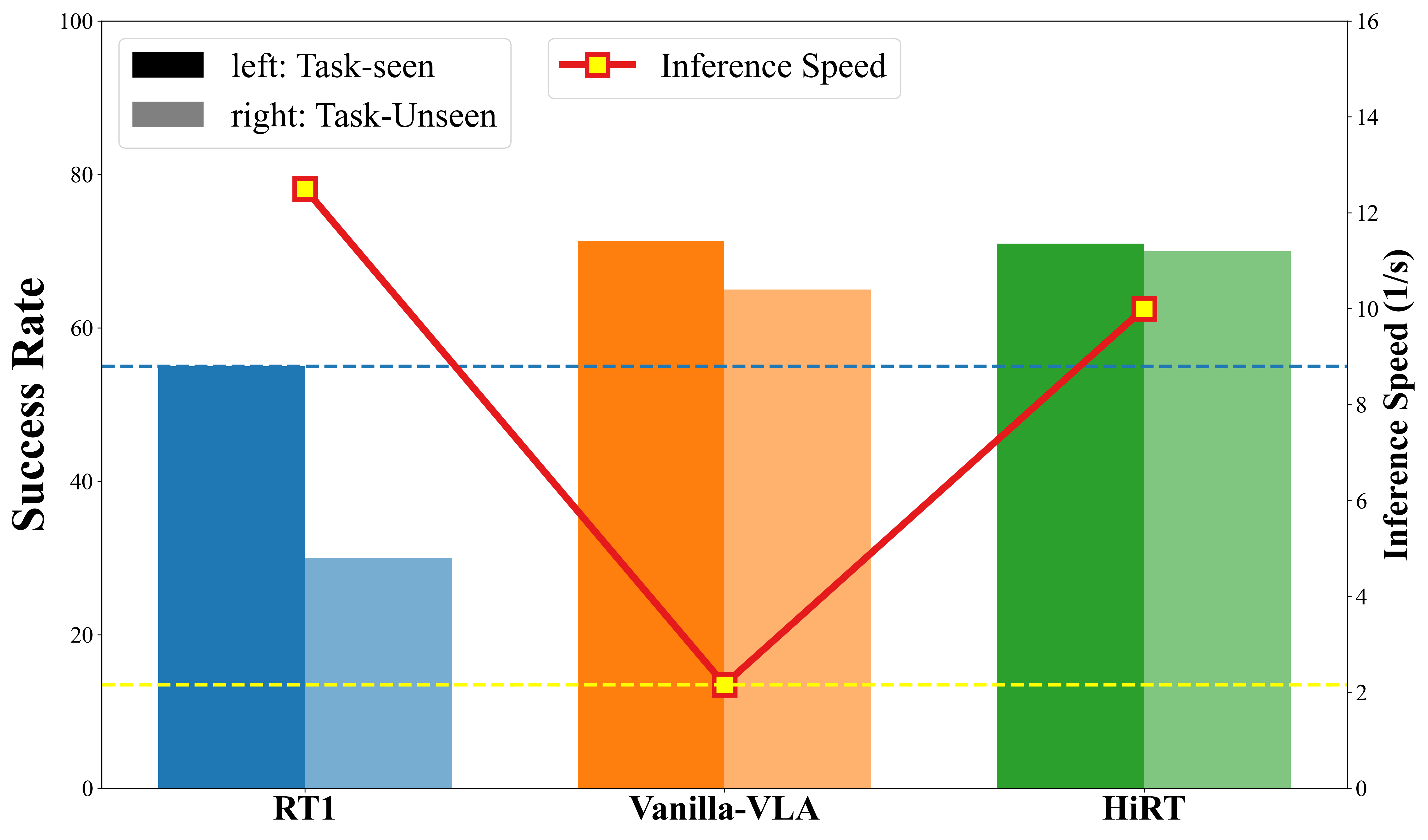}}
\vspace{-2mm}
\caption{Speed and Performance of HiRT with different VLM frequency. Each chart compares the different states of HiRT with Vanilla-VLA and RT-1.
%resulting in the best model performance but the lowest inference speed). 
%The results demonstrate that HiRT can balance performance and inference speed, significantly improving inference speed while retaining the original capabilities of the VLM.
}
    %%在每张图中，将不同状态的HiRT与Vanilla-VLA和RT-1比较，其中$1/f_{VLM}$表示HiRT中VLM的运行间隔（1表示VLM以最大频率执行，此时模型性能最好，但推理速度最低）。
    %结果表明了HiRT可以调整performance和推理速度，保留原本VLM能力的同时很大程度提高推理时的速度。
\label{chart: speed}
\vspace{-7.3mm}
\end{center}
\end{figure}

\textbf{Balancing between Performance and Efficiency.}
%为了充分发挥模型的性能以进行公平的比较，之前的实验中HiRT模型均采用同步运行，即每一步VLM都会更新指令特征缓冲区供上层进行条件化。为了进一步说明HiRT能够兼顾性能和推理速度，我们测试了VLM以不同异步速度运行时模型推理所用的时间和任务成功率。由于仿真需要消耗大量CPU和GPU时间，实验中直接统计策略连续处理100帧数据的平均时间作为推理速度的评估。
%图\ref{chart: speed}中展示了将Vanilla-VLA，RT-1和HiRT以不同VLM频率运行时的测试结果。整体来看，随着VLM运行间隔的增大，HiRT模型的在性能方面会由显著高于Vanilla-VLA逐渐下降到接近RT-1的水平，而推理速度却会由Vanilla-VLA（2.15Hz）逐步提升到高于RT-1。这说明HiRT可以在保持模型泛化能力的同时显著提升推理速度。可以观察到HiRT方法在VLM更新间隔为1时的性能远好于同为同步运行的Vanilla-VLA，而二者拥有几乎相同的推理速度。当HiRT以间隔为2运行VLM时，其性能在略高于RobotVLM的同时推理速度显著提高至3.60Hz，接近原本速度的2倍，当以4间隔运行时，其性能较RobotVLM损失了仅3.8\%，推理速度却提升到原本的3倍（6.53Hz）。随着间隔进一步增大到6，模型性能和推理效率都高于RT1水平。
% To fully leverage the model's performance and ensure a fair comparison, the previous experiments with HiRT utilized synchronous operation, where the VLM updates the instruction feature buffer at every step for conditioning the upper layers. 
To further demonstrate HiRT's ability to balance performance and execution frequency, we evaluate the model's inference speed and task success rate. Results are shown in Figure \ref{chart: speed}. 
Notably, HiRT's performance is comparable with Vanilla-VLA, while its inference speed significantly increases to 9.8Hz, nearly doubling the original speed. This indicates that HiRT can significantly enhance inference speed while maintaining model generalization capabilities. 
\subsection{Performance on real-world dynamic manipulation tasks}\label{dynamicexp}
\begin{figure}[H]
\vspace{-3mm}
\begin{minipage}{.4\linewidth}
% \vspace{-2mm}
\begin{figure}[H]
    \centering
    \includegraphics[width=0.95\linewidth]{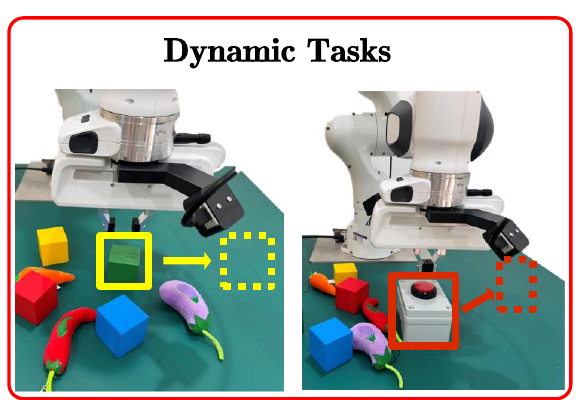}
    \vspace{-1mm}
    \caption{Visualized Dynamic Tasks.}\label{fig: dynamictask}
\end{figure}
\vspace{-3mm}
\end{minipage}
\hspace{1mm}
\begin{minipage}{.55\linewidth}
\vspace{-1mm}
\begin{table}[H]
% \small
\centering
\resizebox{1.0\linewidth}{!}{
\begin{tabular}{cccccc}
\toprule
\textbf{Method}&\textbf{Time/s$\downarrow$}  &\textbf{Seen$\uparrow$} &\textbf{Unseen$\uparrow$}& \textbf{Average$\uparrow$}\\
\midrule
\textbf{Diffusion Policy}& 10.38 & 20.0 & 15.0 & 18.0\\
\textbf{RT-1*}   &          14.22 & 25.0 & 10.0 & 18.0\\
\textbf{Vanilla-VLA}     &   9.25 & 55.0 & 40.0 & 48.0\\
\textbf{HiRT}    &   \textbf{6.18} & \textbf{80.0} & \textbf{70.0}   & \textbf{75.0}\\ 
% \textbf{HiRT-2}    &   \textbf{0.80} & 0.55& 0.76   & 0.72\\ 
% \textbf{HiRT-4}     &   \textbf{0.80} & 0.55& 0.76   & 0.72\\ 
% \textbf{HiRT-6}    &   \textbf{0.80} & 0.55& 0.76   & 0.72\\ 
    \bottomrule
\end{tabular}
}
\vspace{2mm}
\caption{Success rates on real-world dynamic manipulation tasks. With our hierarchical design, HiRT achieves the highest success rate and finishes the task in the least time.}
\vspace{-4mm}
\label{tab: dynamics}
\end{table}

\end{minipage}
\vspace{-6mm}
\end{figure}

%放个数据演示
%放个不同频率模型的完成率表（和上一个图可以并排）
%放点比较demo
% 
%我们在真实世界中的一系列动态PickandPlace任务中测试了具有不同频率和性能的模型。In particular, 我们在图\ref{fig: dynamictask}所示的不同泛化难度下的场景中以恒定速度（1cm/s）移动目标物体来模拟真实生活环境中可能遇到的操作任务。我们将在静态任务上预训练的策略直接用于动态任务的测试，比较了具有不同频率的HiRT与Vanilla-VLA和RT-1的任务完成率，并以直接在终点目标位置放置目标物体的完成时间作为参考。
%如表\ref{tab: dynamics}所示，以VLM间隔为6运行的HiRT在准静态任务的耗时最低，在in-domain场景中的动态action test中由最高的任务完成率，这得益于它能够高频地关注视觉场景的变化。在新场景的动态任务中，HiRT-4取得了最好的表现，透露了它在具有较高运行频率的同时具有较强的泛化能力。整体来说，HiRT方法能够将基于VLM的策略应用在需要高频反应的任务中，这是先前方法所无法做到的。
We evaluate models with varying frequencies and performance levels on a series of dynamic tasks in real-world scenarios. Specifically, as illustrated in Figure \ref{fig: dynamictask}, we simulate realistic operational tasks by moving target objects at a constant speed of 1 cm/s, presenting different levels of generalization. 
%We directly test strategies pre-trained on static tasks, comparing the task completion rates of HiRT at different frequencies with other baselines.

\begin{figure}[t]
    \centering
    \includegraphics[width=1.0\textwidth]{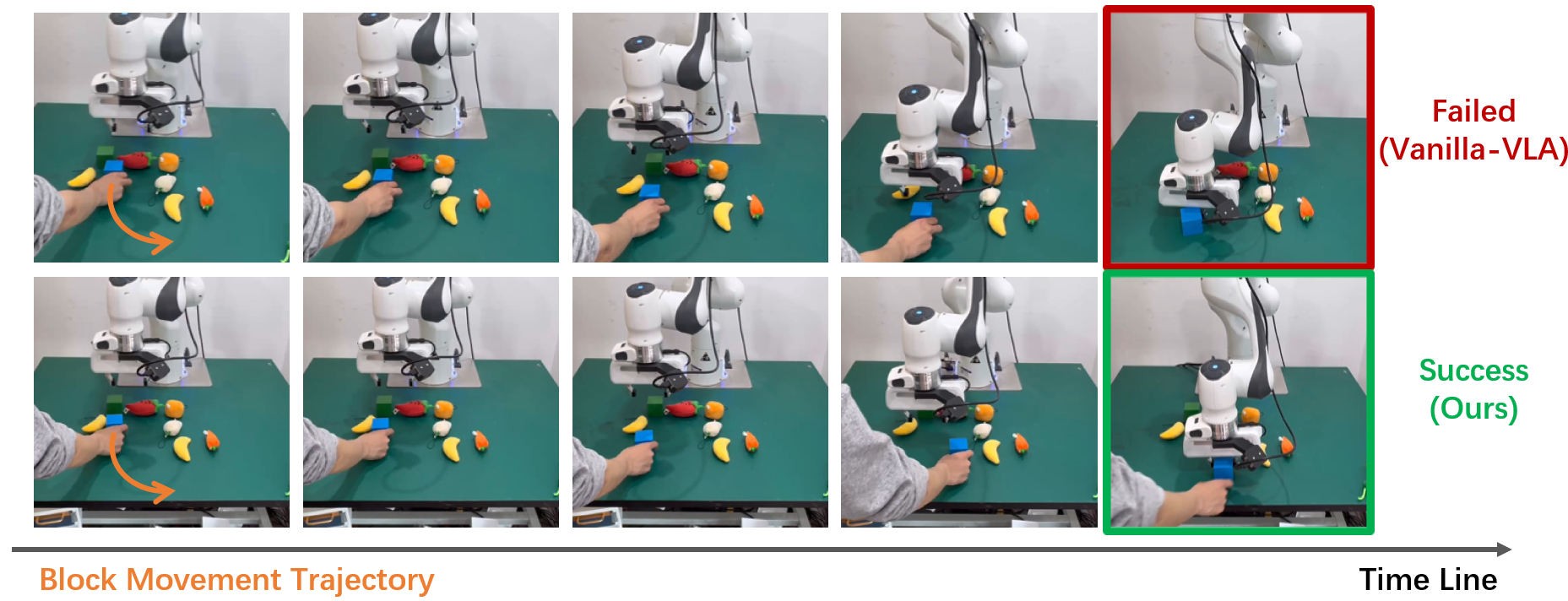}
    % \vspace{0.2mm}
    \vspace{-1mm}
    \caption{Comparisons under dynamics real-world experiments. The blue block is moving along a trajectory on the table. Our method successfully tracks the movement of the blue block and catches it. While the baseline method misses the blue block due to long inference time and high latency.}
    \label{fig: dynamic demo}
    
\end{figure}

Results are shown in Table \ref{tab: dynamics}, where the column of \textit{Time} represents the duration taken by the model to complete the quasi-static task in the same scene without moving objects, serving as a reference of the model's efficiency. HiRT achieves the shortest completion time in quasi-static tasks and the highest task success rate in dynamic action tests within both in-domain and out-domain scenarios, indicating its strong generalization capability while maintaining a high execution speed. Overall, the HiRT approach effectively applies VLM-based methods to various dynamic tasks. A comparision example is visualized in Figure \ref{fig: dynamic demo}.

\subsection{Ablation Study on implementation of HiRT}\label{ablation}
In this experiment, we seek to understand the different components of HiRT. Specifically, we compare the full HiRT with
\textbf{HiRT(-IC)}, which uses the most recent single frame for latent-conditioned-policy, 
% with \textbf{HiRT(-TF)}, which does not use conditioned-transformer in latent-conditioned-policy, 
\textbf{HiRT(-CD)}, which replaces the conditioned-transformer and conditioned MAP (keep FiLM block which has been validated as effective in RT-1~\citep{brohan2022rt}) with the original module.
% \textbf{HiRT(-AS)}, which utilizes the same input of latent-conditioned-policy.
We primarily conduct module ablation in the Franka-Kitchen environment because it allows for rapid testing on generalization capability. More ablations can be found in Appendix \ref{app_meta}.
\begin{table}[htbp]
\centering
\centering
\resizebox{1.0\linewidth}{!}{
\begin{tabular}{c|c|ccccc|c|cc|c}
%   \begin{tabular}{*{1}{>{\centering\arraybackslash}p{2cm}}|*{5}{>{\centering\arraybackslash}p{0.8cm}}|*{1}{>{\centering\arraybackslash}p{1.15cm}}|*{2}{>{\centering\arraybackslash}p{0.8cm}}|*{1}{>{\centering\arraybackslash}p{1.15cm}}}
\toprule
% \textbf{Methods} & \textbf{\begin{tabular}[c]{@{}c@{}}button-\\ press\end{tabular}} & \textbf{\begin{tabular}[c]{@{}c@{}} button\\ topdown\end{tabular}} & \textbf{\begin{tabular}[c]{@{}c@{}}drawer-\\ open\end{tabular}} & \textbf{\begin{tabular}[c]{@{}c@{}}door-\\ open\end{tabular}} & \textbf{\begin{tabular}[c]{@{}c@{}}faucet-\\ close\end{tabular}} & \textbf{\begin{tabular}[c]{@{}c@{}}plate-\\ slide\end{tabular}} & \textbf{\begin{tabular}[c]{@{}c@{}}reach-\\ wall\end{tabular}} & \textbf{\begin{tabular}[c]{@{}c@{}}window-\\ open\end{tabular}} & \textbf{\begin{tabular}[c]{@{}c@{}}window-\\ close\end{tabular}} & \textbf{\begin{tabular}[c]{@{}c@{}}door-\\ lock\end{tabular}}  \\ \midrule
\textbf{Method}&\textbf{Infer Speed}$\uparrow$&\textbf{Knob}&\textbf{Ldoor}&\textbf{Light}&\textbf{Micro}&\textbf{Rdoor}&$\bm{Avg_{seen}}$&\textbf{Light*}&\textbf{Rdoor*}&$\bm{Avg_{new}}$\\\hline
RT-1&20.1&\textbf{90}&\textbf{72}&51&87&19&63.8&1&65&33\\
Vanilla-VLA&4.1&84&43&96&67&77&73.4&\textbf{85}&46&65.5\\
    HiRT-IC&10.1&83&0&93&18&\textbf{100}&58.8&28&66&47\\
    % HiRT-IC-TF&1.97+++&33&33&44&52&86&49.6&32&24&28\\
    HiRT-IC-CD&11.0&87&35&39&6&73&48.0&0&19&9.5\\
    % HiRT-TF&1.97++&70&66&97&83&\textbf{100}&83.2&59&96&75.5\\
HiRT-CD&10.9&79&14&74&64&90&64.2&39&84&61.5\\
% HiRT-AS&89&9&93&\textbf{99}&97&77.4&16&36&26.0\\
\rowcolor[gray]{0.8}\textbf{HiRT}&9.8&84&43&\textbf{99}&\textbf{79}&99&\underline{\textbf{\textit{80.8}}}&52&\textbf{100}&\underline{\textbf{\textit{76.0}}}\\
 \bottomrule
\end{tabular}
}
\vspace{2mm}
\caption{Ablating Components of HiRT. Results with \textit{-IC} reveal that image context is important for good performance. Using the combined conditioning strategy leads to a 20\% increase in success rate.}\label{tab: ablation}
% Results with -IC reveals that image context is important for good performance. Using the original transformer like RT-1 lead to 20\% decrease in success rate, while adding the condition layers can bootstrap performance of the action policy.
\vspace{-6mm}
\end{table}
\vspace{-1mm}

\textbf{Does VLM-Latent Conditioning Improve Multi-Task Performance?}
% 使用VLM-Latent进行条件化是否有助于提高动作策略在多任务上的性能？
%我们测试了latent-conditioned-policy中采用不同条件化策略。如图\ref{tab:ablation}, 可以看到full-HiRT在多任务和新任务上都有最高的任务完成率。去除了Transformer中条件化层（HiRT-CD）会使其下降20\%成功率，说明了条件化对于多任务学习十分重要。值得注意的是，去除整个Transformer的HiRT-TF（即latent-conditioned-policy仅包含FiLM-conditioned-EfficientNet）相较于HiRT-CD有更少的参数和更高的推理速度，但性能却有明显提升，这说明对全部编码层使用条件化有助于动作策略分辨不同的任务。
% We evaluate the impact of different conditioning strategies on the performance of the latent-conditioned policy. 
As shown in Table \ref{tab: ablation}, the full HiRT model achieves the highest task completion rates in both multi-task and new task settings. Removing the extra conditioning layers (HiRT-CD) results in a 20\% decrease in success rate, highlighting the importance of conditioning for multi-task learning. 
%Notably, removing the entire Transformer (HiRT-TF), leaving only the FiLM-conditioned EfficientNet in the latent-conditioned policy, results in fewer parameters and higher inference speed compared to HiRT-CD, yet it shows a significant performance improvement. This indicates that applying conditioning across all encoding layers helps the action policy better distinguish between different tasks.

\textbf{How Does the Latent-Conditioned Policy Perform with Different Visual Inputs?}
% 快速的Latent-Conditioned-Policy在不同的视觉输入下会有如何的表现？
%我们对latent-conditioned-policy输入的视觉信息不同形式进行了比较，包括单张最近时间步的视觉观察(HiRT-IC)和默认的过去6张连续的视觉上下文。如表\ref{tab: ablation}所示，不论采用那种模型结构（e.g. full，-CD，-TF），采用了连续多张的视觉上下文的方法相较于单张图片的方法在多任务上有巨大性能提升，泛化在新的场景的能力也有近30\%的明显增长。尽管以多图上下文作为输入会降低推理速度，但其能够显著提升增强动作策略对场景的认知，利用同样的VLM-Latent输出更准确的动作。
% We compare the performance of the latent-conditioned policy with different forms of visual inputs, including a single recent visual observation (HiRT-IC) and the default setting of the past six images in \mbox{RT-1~\citep{brohan2022rt}}. 
In Table \ref{tab: ablation}, regardless of the model structure (e.g. full, -CD), using multiple consecutive visual inputs significantly outperforms using a single image (results with \textit{-IC}).
% This method results in a substantial performance boost in multi-task settings and nearly a 30\% improvement in generalization to new scenes. 
Although using multiple visual contexts as input can reduce inference speed, it greatly enhances the policy's understanding of the scene, enabling it to generate more accurate actions using the same VLM-Latent output.

\section{Conclusions, Limitations and Future Works}
\label{sec:conclusion}
% We present HiRT, a novel framework to predict future images and generate actions under a joint313
% denoising process. Moreover, PAD can co-train with internet video datasets and extend to other314
% robotic modalities. Both simulated and real-world experiments demonstrated the efficiency of PAD.315
% A limitation of the current method is that we only tested with three types of modalities. Subsequent316
% endeavors could extend this framework to incorporate additional robot-related input data, such as317
% tactile information, which we believe are valuable research directions.
In conclusion, this study addresses the limitations of VLMs in handling complex dynamic tasks due to high computational costs and inference delays. By proposing HiRT, a hierarchical imitation learning framework, we enhance execution speed and multi-task generalization. However, due to data constraints, we have not yet employed HiRT on more complex dynamic tasks, such as grasping high-speed flying objects or rapidly adjusting object postures. This could be an area for our future research with more specific task data and adaptive adjustments to certain modules.
%===============================================================================

% \section{Citations}
% \label{sec:citations}

% 	Citations can be made using either \textbackslash citep\{\} or \textbackslash citet\{\}, depending from the appropriateness. To avoid the citation moving to the next line, it is often a good practice to replace the space before with a tilde (\~{}) character.
% 	Example 1: ``CoRL is the best conference ever~\citep{Gauss1857}.''
% 	Example 2: ``\citet{Lagrange1788} proved, both theoretically and numerically, that CoRL is the best conference ever.''
	
%===============================================================================

%===============================================================================

%===============================================================================

\clearpage
% The acknowledgments are automatically included only in the final and preprint versions of the paper.
% \acknowledgments{}

%===============================================================================

% no \bibliographystyle is required, since the corl style is automatically used.
\bibliography{main}  % .bib

\begin{thebibliography}{63}
\providecommand{\natexlab}[1]{#1}
\providecommand{\url}[1]{\texttt{#1}}
\expandafter\ifx\csname urlstyle\endcsname\relax
  \providecommand{\doi}[1]{doi: #1}\else
  \providecommand{\doi}{doi: \begingroup \urlstyle{rm}\Url}\fi

\bibitem[Brohan et~al.(2023)Brohan, Brown, Carbajal, Chebotar, Chen, Choromanski, Ding, Driess, Dubey, Finn, et~al.]{brohan2023rt}
A.~Brohan, N.~Brown, J.~Carbajal, Y.~Chebotar, X.~Chen, K.~Choromanski, T.~Ding, D.~Driess, A.~Dubey, C.~Finn, et~al.
\newblock Rt-2: Vision-language-action models transfer web knowledge to robotic control.
\newblock \emph{arXiv preprint arXiv:2307.15818}, 2023.

\bibitem[Li et~al.(2023{\natexlab{a}})Li, Liu, Zhang, Yu, Xu, Wu, Cheang, Jing, Zhang, Liu, et~al.]{li2023vision}
X.~Li, M.~Liu, H.~Zhang, C.~Yu, J.~Xu, H.~Wu, C.~Cheang, Y.~Jing, W.~Zhang, H.~Liu, et~al.
\newblock Vision-language foundation models as effective robot imitators.
\newblock \emph{arXiv preprint arXiv:2311.01378}, 2023{\natexlab{a}}.

\bibitem[Li et~al.(2023{\natexlab{b}})Li, Li, Savarese, and Hoi]{li2023blip}
J.~Li, D.~Li, S.~Savarese, and S.~Hoi.
\newblock Blip-2: Bootstrapping language-image pre-training with frozen image encoders and large language models.
\newblock In \emph{International conference on machine learning}, pages 19730--19742. PMLR, 2023{\natexlab{b}}.

\bibitem[Wang et~al.(2022)Wang, Bao, Dong, Bjorck, Peng, Liu, Aggarwal, Mohammed, Singhal, Som, et~al.]{wang2022image}
W.~Wang, H.~Bao, L.~Dong, J.~Bjorck, Z.~Peng, Q.~Liu, K.~Aggarwal, O.~K. Mohammed, S.~Singhal, S.~Som, et~al.
\newblock Image as a foreign language: Beit pretraining for all vision and vision-language tasks.
\newblock \emph{arXiv preprint arXiv:2208.10442}, 2022.

\bibitem[Dai et~al.(2024)Dai, Li, Li, Tiong, Zhao, Wang, Li, Fung, and Hoi]{dai2024instructblip}
W.~Dai, J.~Li, D.~Li, A.~M.~H. Tiong, J.~Zhao, W.~Wang, B.~Li, P.~N. Fung, and S.~Hoi.
\newblock Instructblip: Towards general-purpose vision-language models with instruction tuning.
\newblock \emph{Advances in Neural Information Processing Systems}, 36, 2024.

\bibitem[Driess et~al.(2023)Driess, Xia, Sajjadi, Lynch, Chowdhery, Ichter, Wahid, Tompson, Vuong, Yu, et~al.]{driess2023palm}
D.~Driess, F.~Xia, M.~S. Sajjadi, C.~Lynch, A.~Chowdhery, B.~Ichter, A.~Wahid, J.~Tompson, Q.~Vuong, T.~Yu, et~al.
\newblock Palm-e: An embodied multimodal language model.
\newblock \emph{arXiv preprint arXiv:2303.03378}, 2023.

\bibitem[Ha and Song(2022)]{ha2022flingbot}
H.~Ha and S.~Song.
\newblock Flingbot: The unreasonable effectiveness of dynamic manipulation for cloth unfolding.
\newblock In \emph{Conference on Robot Learning}, pages 24--33. PMLR, 2022.

\bibitem[Saxena et~al.(2024)Saxena, Sharma, and Kroemer]{saxena2024mrest}
S.~Saxena, M.~Sharma, and O.~Kroemer.
\newblock Mrest: Multi-resolution sensing for real-time control with vision-language models.
\newblock \emph{arXiv preprint arXiv:2401.14502}, 2024.

\bibitem[Wason and Evans(1974)]{wason1974dual}
P.~C. Wason and J.~S.~B. Evans.
\newblock Dual processes in reasoning?
\newblock \emph{Cognition}, 3\penalty0 (2):\penalty0 141--154, 1974.

\bibitem[MacMahon et~al.(2006)MacMahon, Stankiewicz, and Kuipers]{macmahon2006walk}
M.~MacMahon, B.~Stankiewicz, and B.~Kuipers.
\newblock Walk the talk: Connecting language, knowledge, and action in route instructions.
\newblock \emph{Def}, 2\penalty0 (6):\penalty0 4, 2006.

\bibitem[Tellex et~al.(2011)Tellex, Kollar, Dickerson, Walter, Banerjee, Teller, and Roy]{tellex2011understanding}
S.~Tellex, T.~Kollar, S.~Dickerson, M.~Walter, A.~Banerjee, S.~Teller, and N.~Roy.
\newblock Understanding natural language commands for robotic navigation and mobile manipulation.
\newblock In \emph{Proceedings of the AAAI Conference on Artificial Intelligence}, volume~25, pages 1507--1514, 2011.

\bibitem[Tellex et~al.(2020)Tellex, Gopalan, Kress-Gazit, and Matuszek]{tellex2020robots}
S.~Tellex, N.~Gopalan, H.~Kress-Gazit, and C.~Matuszek.
\newblock Robots that use language.
\newblock \emph{Annual Review of Control, Robotics, and Autonomous Systems}, 3:\penalty0 25--55, 2020.

\bibitem[Brohan et~al.(2022)Brohan, Brown, Carbajal, Chebotar, Dabis, Finn, Gopalakrishnan, Hausman, Herzog, Hsu, et~al.]{brohan2022rt}
A.~Brohan, N.~Brown, J.~Carbajal, Y.~Chebotar, J.~Dabis, C.~Finn, K.~Gopalakrishnan, K.~Hausman, A.~Herzog, J.~Hsu, et~al.
\newblock Rt-1: Robotics transformer for real-world control at scale.
\newblock \emph{arXiv preprint arXiv:2212.06817}, 2022.

\bibitem[Stone et~al.(2023)Stone, Xiao, Lu, Gopalakrishnan, Lee, Vuong, Wohlhart, Kirmani, Zitkovich, Xia, et~al.]{stone2023open}
A.~Stone, T.~Xiao, Y.~Lu, K.~Gopalakrishnan, K.-H. Lee, Q.~Vuong, P.~Wohlhart, S.~Kirmani, B.~Zitkovich, F.~Xia, et~al.
\newblock Open-world object manipulation using pre-trained vision-language models.
\newblock \emph{arXiv preprint arXiv:2303.00905}, 2023.

\bibitem[Jang et~al.(2022)Jang, Irpan, Khansari, Kappler, Ebert, Lynch, Levine, and Finn]{jang2022bc}
E.~Jang, A.~Irpan, M.~Khansari, D.~Kappler, F.~Ebert, C.~Lynch, S.~Levine, and C.~Finn.
\newblock Bc-z: Zero-shot task generalization with robotic imitation learning.
\newblock In \emph{Conference on Robot Learning}, pages 991--1002. PMLR, 2022.

\bibitem[Thomason et~al.(2020)Thomason, Padmakumar, Sinapov, Walker, Jiang, Yedidsion, Hart, Stone, and Mooney]{thomason2020jointly}
J.~Thomason, A.~Padmakumar, J.~Sinapov, N.~Walker, Y.~Jiang, H.~Yedidsion, J.~Hart, P.~Stone, and R.~Mooney.
\newblock Jointly improving parsing and perception for natural language commands through human-robot dialog.
\newblock \emph{Journal of Artificial Intelligence Research}, 67:\penalty0 327--374, 2020.

\bibitem[Li et~al.(2023)Li, Hammoud, Itani, Khizbullin, and Ghanem]{li2023camel}
G.~Li, H.~A. A.~K. Hammoud, H.~Itani, D.~Khizbullin, and B.~Ghanem.
\newblock Camel: Communicative agents for" mind" exploration of large scale language model society.
\newblock 2023.

\bibitem[Ma et~al.(2023)Ma, Kumar, Zhang, Bastani, and Jayaraman]{ma2023liv}
Y.~J. Ma, V.~Kumar, A.~Zhang, O.~Bastani, and D.~Jayaraman.
\newblock Liv: Language-image representations and rewards for robotic control.
\newblock In \emph{International Conference on Machine Learning}, pages 23301--23320. PMLR, 2023.

\bibitem[Nair et~al.(2022{\natexlab{a}})Nair, Rajeswaran, Kumar, Finn, and Gupta]{nair2022r3m}
S.~Nair, A.~Rajeswaran, V.~Kumar, C.~Finn, and A.~Gupta.
\newblock R3m: A universal visual representation for robot manipulation.
\newblock \emph{arXiv preprint arXiv:2203.12601}, 2022{\natexlab{a}}.

\bibitem[Nair et~al.(2022{\natexlab{b}})Nair, Mitchell, Chen, Savarese, Finn, et~al.]{nair2022learning}
S.~Nair, E.~Mitchell, K.~Chen, S.~Savarese, C.~Finn, et~al.
\newblock Learning language-conditioned robot behavior from offline data and crowd-sourced annotation.
\newblock In \emph{Conference on Robot Learning}, pages 1303--1315. PMLR, 2022{\natexlab{b}}.

\bibitem[Luketina et~al.(2019)Luketina, Nardelli, Farquhar, Foerster, Andreas, Grefenstette, Whiteson, and Rockt{\"a}schel]{luketina2019survey}
J.~Luketina, N.~Nardelli, G.~Farquhar, J.~Foerster, J.~Andreas, E.~Grefenstette, S.~Whiteson, and T.~Rockt{\"a}schel.
\newblock A survey of reinforcement learning informed by natural language.
\newblock \emph{IJCAI}, 2019.

\bibitem[Misra et~al.(2017)Misra, Langford, and Artzi]{misra2017mapping}
D.~Misra, J.~Langford, and Y.~Artzi.
\newblock Mapping instructions and visual observations to actions with reinforcement learning.
\newblock \emph{arXiv preprint arXiv:1704.08795}, 2017.

\bibitem[Jiang et~al.(2019)Jiang, Gu, Murphy, and Finn]{jiang2019language}
Y.~Jiang, S.~S. Gu, K.~P. Murphy, and C.~Finn.
\newblock Language as an abstraction for hierarchical deep reinforcement learning.
\newblock \emph{Advances in Neural Information Processing Systems}, 32, 2019.

\bibitem[Goyal et~al.(2021)Goyal, Niekum, and Mooney]{goyal2021pixl2r}
P.~Goyal, S.~Niekum, and R.~Mooney.
\newblock Pixl2r: Guiding reinforcement learning using natural language by mapping pixels to rewards.
\newblock In \emph{Conference on Robot Learning}, pages 485--497. PMLR, 2021.

\bibitem[Oh et~al.(2017)Oh, Singh, Lee, and Kohli]{oh2017zero}
J.~Oh, S.~Singh, H.~Lee, and P.~Kohli.
\newblock Zero-shot task generalization with multi-task deep reinforcement learning.
\newblock In \emph{International Conference on Machine Learning}, pages 2661--2670. PMLR, 2017.

\bibitem[Andreas et~al.(2017)Andreas, Klein, and Levine]{andreas2017modular}
J.~Andreas, D.~Klein, and S.~Levine.
\newblock Modular multitask reinforcement learning with policy sketches.
\newblock In \emph{International conference on machine learning}, pages 166--175. PMLR, 2017.

\bibitem[Ahn et~al.(2022)Ahn, Brohan, Brown, Chebotar, Cortes, David, Finn, Fu, Gopalakrishnan, Hausman, et~al.]{ahn2022can}
M.~Ahn, A.~Brohan, N.~Brown, Y.~Chebotar, O.~Cortes, B.~David, C.~Finn, C.~Fu, K.~Gopalakrishnan, K.~Hausman, et~al.
\newblock Do as i can, not as i say: Grounding language in robotic affordances.
\newblock \emph{arXiv preprint arXiv:2204.01691}, 2022.

\bibitem[Crispino et~al.(2023)Crispino, Montgomery, Zeng, Song, and Wang]{crispino2023agent}
N.~Crispino, K.~Montgomery, F.~Zeng, D.~Song, and C.~Wang.
\newblock Agent instructs large language models to be general zero-shot reasoners.
\newblock \emph{arXiv preprint arXiv:2310.03710}, 2023.

\bibitem[Stepputtis et~al.(2020)Stepputtis, Campbell, Phielipp, Lee, Baral, and Ben~Amor]{stepputtis2020language}
S.~Stepputtis, J.~Campbell, M.~Phielipp, S.~Lee, C.~Baral, and H.~Ben~Amor.
\newblock Language-conditioned imitation learning for robot manipulation tasks.
\newblock \emph{Advances in Neural Information Processing Systems}, 33:\penalty0 13139--13150, 2020.

\bibitem[Shridhar et~al.(2022)Shridhar, Manuelli, and Fox]{shridhar2022cliport}
M.~Shridhar, L.~Manuelli, and D.~Fox.
\newblock Cliport: What and where pathways for robotic manipulation.
\newblock In \emph{Conference on robot learning}, pages 894--906. PMLR, 2022.

\bibitem[Mei et~al.(2016)Mei, Bansal, and Walter]{mei2016listen}
H.~Mei, M.~Bansal, and M.~Walter.
\newblock Listen, attend, and walk: Neural mapping of navigational instructions to action sequences.
\newblock In \emph{Proceedings of the AAAI Conference on Artificial Intelligence}, volume~30, 2016.

\bibitem[Wang et~al.(2023)Wang, Zhang, Chen, and Sreenath]{wang2023prompt}
Y.-J. Wang, B.~Zhang, J.~Chen, and K.~Sreenath.
\newblock Prompt a robot to walk with large language models.
\newblock \emph{arXiv preprint arXiv:2309.09969}, 2023.

\bibitem[Zeng et~al.(2022)Zeng, Attarian, Ichter, Choromanski, Wong, Welker, Tombari, Purohit, Ryoo, Sindhwani, et~al.]{zeng2022socratic}
A.~Zeng, M.~Attarian, B.~Ichter, K.~Choromanski, A.~Wong, S.~Welker, F.~Tombari, A.~Purohit, M.~Ryoo, V.~Sindhwani, et~al.
\newblock Socratic models: Composing zero-shot multimodal reasoning with language.
\newblock \emph{arXiv preprint arXiv:2204.00598}, 2022.

\bibitem[Shah et~al.(2023)Shah, Osi{\'n}ski, Levine, et~al.]{shah2023lm}
D.~Shah, B.~Osi{\'n}ski, S.~Levine, et~al.
\newblock Lm-nav: Robotic navigation with large pre-trained models of language, vision, and action.
\newblock In \emph{Conference on robot learning}, pages 492--504. PMLR, 2023.

\bibitem[Huang et~al.(2022)Huang, Xia, Xiao, Chan, Liang, Florence, Zeng, Tompson, Mordatch, Chebotar, et~al.]{huang2022inner}
W.~Huang, F.~Xia, T.~Xiao, H.~Chan, J.~Liang, P.~Florence, A.~Zeng, J.~Tompson, I.~Mordatch, Y.~Chebotar, et~al.
\newblock Inner monologue: Embodied reasoning through planning with language models.
\newblock \emph{arXiv preprint arXiv:2207.05608}, 2022.

\bibitem[Huang et~al.(2023)Huang, Mees, Zeng, and Burgard]{huang2023visual}
C.~Huang, O.~Mees, A.~Zeng, and W.~Burgard.
\newblock Visual language maps for robot navigation.
\newblock In \emph{2023 IEEE International Conference on Robotics and Automation (ICRA)}, pages 10608--10615. IEEE, 2023.

\bibitem[Song et~al.(2023)Song, Wu, Washington, Sadler, Chao, and Su]{song2023llm}
C.~H. Song, J.~Wu, C.~Washington, B.~M. Sadler, W.-L. Chao, and Y.~Su.
\newblock Llm-planner: Few-shot grounded planning for embodied agents with large language models.
\newblock In \emph{Proceedings of the IEEE/CVF International Conference on Computer Vision}, pages 2998--3009, 2023.

\bibitem[Liu et~al.(2023)Liu, Jiang, Zhang, Liu, Zhang, Biswas, and Stone]{liu2023llm+}
B.~Liu, Y.~Jiang, X.~Zhang, Q.~Liu, S.~Zhang, J.~Biswas, and P.~Stone.
\newblock Llm+ p: Empowering large language models with optimal planning proficiency.
\newblock \emph{arXiv preprint arXiv:2304.11477}, 2023.

\bibitem[Mu et~al.(2024)Mu, Zhang, Hu, Wang, Ding, Jin, Wang, Dai, Qiao, and Luo]{mu2024embodiedgpt}
Y.~Mu, Q.~Zhang, M.~Hu, W.~Wang, M.~Ding, J.~Jin, B.~Wang, J.~Dai, Y.~Qiao, and P.~Luo.
\newblock Embodiedgpt: Vision-language pre-training via embodied chain of thought.
\newblock \emph{Advances in Neural Information Processing Systems}, 36, 2024.

\bibitem[Wu et~al.(2023)Wu, Jing, Cheang, Chen, Xu, Li, Liu, Li, and Kong]{wu2023unleashing}
H.~Wu, Y.~Jing, C.~Cheang, G.~Chen, J.~Xu, X.~Li, M.~Liu, H.~Li, and T.~Kong.
\newblock Unleashing large-scale video generative pre-training for visual robot manipulation.
\newblock \emph{arXiv preprint arXiv:2312.13139}, 2023.

\bibitem[Bahl et~al.(2023)Bahl, Mendonca, Chen, Jain, and Pathak]{bahl2023affordances}
S.~Bahl, R.~Mendonca, L.~Chen, U.~Jain, and D.~Pathak.
\newblock Affordances from human videos as a versatile representation for robotics.
\newblock In \emph{Proceedings of the IEEE/CVF Conference on Computer Vision and Pattern Recognition}, pages 13778--13790, 2023.

\bibitem[Belkhale et~al.(2024)Belkhale, Ding, Xiao, Sermanet, Vuong, Tompson, Chebotar, Dwibedi, and Sadigh]{belkhale2024rt}
S.~Belkhale, T.~Ding, T.~Xiao, P.~Sermanet, Q.~Vuong, J.~Tompson, Y.~Chebotar, D.~Dwibedi, and D.~Sadigh.
\newblock Rt-h: Action hierarchies using language.
\newblock \emph{arXiv preprint arXiv:2403.01823}, 2024.

\bibitem[Gu et~al.(2023)Gu, Kirmani, Wohlhart, Lu, Arenas, Rao, Yu, Fu, Gopalakrishnan, Xu, et~al.]{gu2023robotic}
J.~Gu, S.~Kirmani, P.~Wohlhart, Y.~Lu, M.~G. Arenas, K.~Rao, W.~Yu, C.~Fu, K.~Gopalakrishnan, Z.~Xu, et~al.
\newblock Robotic task generalization via hindsight trajectory sketches.
\newblock In \emph{The Twelfth International Conference on Learning Representations}, 2023.

\bibitem[Shah et~al.(2023)Shah, Sridhar, Dashora, Stachowicz, Black, Hirose, and Levine]{shah2023vint}
D.~Shah, A.~Sridhar, N.~Dashora, K.~Stachowicz, K.~Black, N.~Hirose, and S.~Levine.
\newblock Vint: A foundation model for visual navigation.
\newblock \emph{arXiv preprint arXiv:2306.14846}, 2023.

\bibitem[Hu et~al.(2023)Hu, Xie, Jain, Francis, Patrikar, Keetha, Kim, Xie, Zhang, Zhao, et~al.]{hu2023toward}
Y.~Hu, Q.~Xie, V.~Jain, J.~Francis, J.~Patrikar, N.~Keetha, S.~Kim, Y.~Xie, T.~Zhang, Z.~Zhao, et~al.
\newblock Toward general-purpose robots via foundation models: A survey and meta-analysis.
\newblock \emph{arXiv preprint arXiv:2312.08782}, 2023.

\bibitem[Du et~al.(2023)Du, Yang, Florence, Xia, Wahid, Ichter, Sermanet, Yu, Abbeel, Tenenbaum, et~al.]{du2023video}
Y.~Du, M.~Yang, P.~Florence, F.~Xia, A.~Wahid, B.~Ichter, P.~Sermanet, T.~Yu, P.~Abbeel, J.~B. Tenenbaum, et~al.
\newblock Video language planning.
\newblock \emph{arXiv preprint arXiv:2310.10625}, 2023.

\bibitem[Abeyruwan et~al.(2023)Abeyruwan, Bewley, Boffi, Choromanski, D’Ambrosio, Jain, Sanketi, Shankar, Sindhwani, Singh, et~al.]{abeyruwan2023agile}
S.~Abeyruwan, A.~Bewley, N.~M. Boffi, K.~M. Choromanski, D.~B. D’Ambrosio, D.~Jain, P.~R. Sanketi, A.~Shankar, V.~Sindhwani, S.~Singh, et~al.
\newblock Agile catching with whole-body mpc and blackbox policy learning.
\newblock In \emph{Learning for Dynamics and Control Conference}, pages 851--863. PMLR, 2023.

\bibitem[Guo et~al.(2023)Guo, Wang, Zha, Jiang, and Chen]{guo2023doremi}
Y.~Guo, Y.-J. Wang, L.~Zha, Z.~Jiang, and J.~Chen.
\newblock Doremi: Grounding language model by detecting and recovering from plan-execution misalignment.
\newblock \emph{arXiv preprint arXiv:2307.00329}, 2023.

\bibitem[Lin et~al.(2023)Lin, Agia, Migimatsu, Pavone, and Bohg]{lin2023text2motion}
K.~Lin, C.~Agia, T.~Migimatsu, M.~Pavone, and J.~Bohg.
\newblock Text2motion: From natural language instructions to feasible plans.
\newblock \emph{Autonomous Robots}, 47\penalty0 (8):\penalty0 1345--1365, 2023.

\bibitem[Singh et~al.(2023)Singh, Blukis, Mousavian, Goyal, Xu, Tremblay, Fox, Thomason, and Garg]{singh2023progprompt}
I.~Singh, V.~Blukis, A.~Mousavian, A.~Goyal, D.~Xu, J.~Tremblay, D.~Fox, J.~Thomason, and A.~Garg.
\newblock Progprompt: Generating situated robot task plans using large language models.
\newblock In \emph{2023 IEEE International Conference on Robotics and Automation (ICRA)}, pages 11523--11530. IEEE, 2023.

\bibitem[Liang et~al.(2023)Liang, Huang, Xia, Xu, Hausman, Ichter, Florence, and Zeng]{liang2023code}
J.~Liang, W.~Huang, F.~Xia, P.~Xu, K.~Hausman, B.~Ichter, P.~Florence, and A.~Zeng.
\newblock Code as policies: Language model programs for embodied control.
\newblock In \emph{2023 IEEE International Conference on Robotics and Automation (ICRA)}, pages 9493--9500. IEEE, 2023.

\bibitem[Wang et~al.(2023)Wang, Ling, Yuan, Shridhar, Bao, Qin, Wang, Xu, and Wang]{wang2023gensim}
L.~Wang, Y.~Ling, Z.~Yuan, M.~Shridhar, C.~Bao, Y.~Qin, B.~Wang, H.~Xu, and X.~Wang.
\newblock Gensim: Generating robotic simulation tasks via large language models.
\newblock \emph{arXiv preprint arXiv:2310.01361}, 2023.

\bibitem[Rana et~al.(2023)Rana, Haviland, Garg, Abou-Chakra, Reid, and Suenderhauf]{rana2023sayplan}
K.~Rana, J.~Haviland, S.~Garg, J.~Abou-Chakra, I.~Reid, and N.~Suenderhauf.
\newblock Sayplan: Grounding large language models using 3d scene graphs for scalable task planning.
\newblock \emph{arXiv preprint arXiv:2307.06135}, 2023.

\bibitem[Huang et~al.(2023)Huang, Wang, Zhang, Li, Wu, and Fei-Fei]{huang2023voxposer}
W.~Huang, C.~Wang, R.~Zhang, Y.~Li, J.~Wu, and L.~Fei-Fei.
\newblock Voxposer: Composable 3d value maps for robotic manipulation with language models.
\newblock \emph{arXiv preprint arXiv:2307.05973}, 2023.

\bibitem[Tang et~al.(2023)Tang, Yu, Tan, Zen, Faust, and Harada]{tang2023saytap}
Y.~Tang, W.~Yu, J.~Tan, H.~Zen, A.~Faust, and T.~Harada.
\newblock Saytap: Language to quadrupedal locomotion.
\newblock \emph{arXiv preprint arXiv:2306.07580}, 2023.

\bibitem[Dosovitskiy et~al.(2020)Dosovitskiy, Beyer, Kolesnikov, Weissenborn, Zhai, Unterthiner, Dehghani, Minderer, Heigold, Gelly, et~al.]{dosovitskiy2020image}
A.~Dosovitskiy, L.~Beyer, A.~Kolesnikov, D.~Weissenborn, X.~Zhai, T.~Unterthiner, M.~Dehghani, M.~Minderer, G.~Heigold, S.~Gelly, et~al.
\newblock An image is worth 16x16 words: Transformers for image recognition at scale.
\newblock \emph{arXiv preprint arXiv:2010.11929}, 2020.

\bibitem[Touvron et~al.(2023)Touvron, Lavril, Izacard, Martinet, Lachaux, Lacroix, Rozi{\`e}re, Goyal, Hambro, Azhar, et~al.]{touvron2023llama}
H.~Touvron, T.~Lavril, G.~Izacard, X.~Martinet, M.-A. Lachaux, T.~Lacroix, B.~Rozi{\`e}re, N.~Goyal, E.~Hambro, F.~Azhar, et~al.
\newblock Llama: Open and efficient foundation language models.
\newblock \emph{arXiv preprint arXiv:2302.13971}, 2023.

\bibitem[Lee et~al.(2019)Lee, Lee, Kim, Kosiorek, Choi, and Teh]{pmlr-v97-lee19d}
J.~Lee, Y.~Lee, J.~Kim, A.~Kosiorek, S.~Choi, and Y.~W. Teh.
\newblock Set transformer: A framework for attention-based permutation-invariant neural networks.
\newblock In K.~Chaudhuri and R.~Salakhutdinov, editors, \emph{Proceedings of the 36th International Conference on Machine Learning}, volume~97 of \emph{Proceedings of Machine Learning Research}, pages 3744--3753. PMLR, 09--15 Jun 2019.
\newblock URL \url{https://proceedings.mlr.press/v97/lee19d.html}.

\bibitem[Tan and Le(2019)]{tan2019efficientnet}
M.~Tan and Q.~Le.
\newblock Efficientnet: Rethinking model scaling for convolutional neural networks.
\newblock In \emph{International conference on machine learning}, pages 6105--6114. PMLR, 2019.

\bibitem[Hu et~al.(2021)Hu, Shen, Wallis, Allen-Zhu, Li, Wang, Wang, and Chen]{hu2021lora}
E.~J. Hu, Y.~Shen, P.~Wallis, Z.~Allen-Zhu, Y.~Li, S.~Wang, L.~Wang, and W.~Chen.
\newblock Lora: Low-rank adaptation of large language models.
\newblock \emph{arXiv preprint arXiv:2106.09685}, 2021.

\bibitem[Yu et~al.(2020)Yu, Quillen, He, Julian, Hausman, Finn, and Levine]{yu2020meta}
T.~Yu, D.~Quillen, Z.~He, R.~Julian, K.~Hausman, C.~Finn, and S.~Levine.
\newblock Meta-world: A benchmark and evaluation for multi-task and meta reinforcement learning.
\newblock In \emph{Conference on robot learning}, pages 1094--1100. PMLR, 2020.

\bibitem[Gupta et~al.(2019)Gupta, Kumar, Lynch, Levine, and Hausman]{gupta2019relay}
A.~Gupta, V.~Kumar, C.~Lynch, S.~Levine, and K.~Hausman.
\newblock Relay policy learning: Solving long-horizon tasks via imitation and reinforcement learning.
\newblock \emph{arXiv preprint arXiv:1910.11956}, 2019.

\bibitem[Chi et~al.(2023)Chi, Feng, Du, Xu, Cousineau, Burchfiel, and Song]{chi2023diffusionpolicy}
C.~Chi, S.~Feng, Y.~Du, Z.~Xu, E.~Cousineau, B.~Burchfiel, and S.~Song.
\newblock Diffusion policy: Visuomotor policy learning via action diffusion.
\newblock In \emph{Proceedings of Robotics: Science and Systems (RSS)}, 2023.

\end{thebibliography}
\newpage
\appendix
\section{Appendix}
\subsection{Implementation Details}\label{app_model}
During implementation, we take use of pretrained EfficientNet-B3 ~\citep{tan2019efficientnet} and ViT-B/16~\citep{dosovitskiy2020image} for the vision encoder of low-level policy, which have been pretrained on large vision data. In training, we insert adapter layers (LoRA layers) throughout the entire InstructBLIP model, including the ViT, Qformer, and LLaMA. In the simulation results, the low-level policy utilize the former CNN architecture, while in the real-world results, the transformer based ViT architecture is employed. For simulation, the fast policy mainly contains the pretrained EfficientNet-B3 vision encoder and the FiLM layers, with totally about 35M parameters. For real world, the fast policy mainly contains the pretrained ViT-B/16 and the cross attention layers, with 150M parameters. 
\subsection{More Details of Experiment Setup}\label{app_setup}
\subsubsection{About Choice of Design for Test Scenarios} 
In the Metaworld simulation environment, we randomly initialize the positions of objects and the robotic arm during testing to assess the method's robustness to object positions in a multi-task setting. In the Franka-kitchen environment, we randomize the relative positions between the robotic arm and the operating platform for each test. Additionally, we significantly alter the color scheme of the scene to evaluate whether the method could complete specific tasks in a visual background that differs greatly from the training data. These two simulation environments are widely used in many studies to validate the fundamental generalization capabilities of models, particularly their robustness to changes in scenes and object positions. 

In our real-world scenarios, the training data only includes situations with a few objects placed on a tabletop. During testing, we not only randomize the positions of the objects and the robotic arm but also place many other objects on the table to introduce distractions. Furthermore, we also test whether the model can grasp entirely new objects it has never seen before to verify its semantic grounding capabilities. Additionally, it's worth mentioning that HiRT's input does not include state information, so all generalization capabilities come from visual and language information.
\subsubsection{About evaluation of generalization capability}

Our tests primarily focus on real-world experiments to validate the semantic generalization capability of our method. In the Metaworld and Franka-kitchen simulation environments, we mainly evaluate the method's generalization to different positions and visual scenes. In the real-world setting, we test whether the model can complete tasks despite the introduction of more distracting objects, a broader range of positional variations, diverse backgrounds, and entirely new objects, e.g. different shapes of vegetables, an arrow-shaped paper, unseen vegetables, toy pizza and blocks with unseen color.
\subsubsection{About data collection details}

For the simulation environment data, we follow the setups of Metaworld and Franka-kitchen by using scripted policies to collect action trajectories for different tasks. In the real world, our data collection is carried out using both manual and scripted methods. 

Specifically, for tasks such as grasping two types of toy fruits (carrot and eggplant), opening drawers, and routing cable tasks, we collect demonstrations manually using a remote operation joystick, ensuring that the target objects are roughly evenly distributed in the field of view. For grasping blocks of different colors, we use scripted policies. We fix four placement positions for the blocks and randomly initialize the robotic arm's position to collect trajectories for these tasks. ( Although the block positions are fixed, we find that HiRT could also correctly grasp blocks placed in novel locations not encountered during training.) 
\subsection{More Ablation}\label{app_meta}

\textbf{How Does Random Sampling in VLM Visual Inputs Affect Model Performance?}

\begin{wraptable}{htbp}{0.7\textwidth}
    \centering
     \begin{tabular}{cc|ccc}
\toprule
&VLM-step&Frequency$\uparrow$&$\bm{Avg_{seen}}$&$\bm{Avg_{new}}$\\\hline
HiRT-AS&1&1.97Hz&96.7&77.2\\
HiRT&1&1.97Hz&94.2&76.0\\
HiRT-AS&6&13.42Hz&52.2&50.0\\
HiRT&6&13.42Hz&63.4&52.5\\
 \bottomrule
\end{tabular}
    \caption{Importance of Asynchronous Sampling.} \label{tab: as}
    \vspace{-2mm}
\end{wraptable}

To determine the impact of sampling images for VLM inputs, we compared the performance of HiRT with HiRT-AS under VLM settings of interval 1 and interval 6. As shown in Table \ref{tab: as}, when VLM synchronizes with the action policy at every step during testing, HiRT-AS slightly outperforms HiRT. This is expected since HiRT-AS uses the most recently updated latent at each step during training. However, when VLM operates with an interval of 6 steps for inference, HiRT-AS shows nearly a 10\% decrease in success rate compared to the original method, indicating that random sampling helps HiRT maintain better generalization performance during asynchronous operation.
% VLM视觉输入中的随机采样如何影响模型的表现？
% 为了确定VLM输入的采样图片的影响，我们比较了HiRT与HiRT-AS在VLM以间隔1和间隔6设置下的表现。如表\ref{tab: as}所示，在测试时VLM在每一步都与动作策略同步运行时，HiRT-AS的性能略高于HiRT，这是符合预期因为训练时HiRT-AS在每一步都会使用当前步最近更新的Latent而在测试时可能遇到多步之前的Latent。然而，在VLM以间隔6运行进行推理时，HiRT-AS相较于原本方法有近\10%的成功率下降，透露出随机采样有助于HiRT在异步运行时保持较好的泛化能力。
\end{document}